\pdfoutput=1

\documentclass[11pt]{article}

\usepackage{acl}

\usepackage{times}
\usepackage{latexsym}

\usepackage[T1]{fontenc}

\usepackage[utf8]{inputenc}

\usepackage{microtype}

\usepackage{inconsolata}

\usepackage{graphicx}

\usepackage{amsmath} 
\usepackage{amssymb}  
\usepackage{booktabs}    
\usepackage{multirow}    
\usepackage{array}       
\usepackage{makecell}    
\usepackage{adjustbox}   

\usepackage{caption}
\usepackage{enumitem}

\usepackage{algorithm}
\usepackage{algpseudocode}

\title{Towards Order Fairness: Mitigating LLMs Order Sensitivity through \\ Dual Group Advantage Optimization}

\author{
  \textbf{Xu Chu\textsuperscript{†}},
  \textbf{Guanyu Wang\textsuperscript{†}},
  \textbf{Zhijie Tan},
  \textbf{Xinrong Chen},
  \textbf{Ziyu Li},
  \textbf{Tong Mo},
  \textbf{Weiping Li\textsuperscript{*}}
\\[0.5em]
  School of Software and Microelectronics, Peking University, Beijing, China
\\
\texttt{\{chuxu, wgy2023, besttangent, chenxinrong23, liziyu\_lizy \}@stu.pku.edu.cn}, 
\\
\texttt{\{motong, wpli\}@ss.pku.edu.cn}
\\[0.5em]
  {
    \textsuperscript{†}Equal contribution
    \hspace{2em}
    \textsuperscript{*}Corresponding author
  }
}

\begin{document}
\maketitle
\begin{abstract}
    Large Language Models (LLMs) suffer from order bias, where their performance is affected by the arrangement order of input elements. This unfairness limits the model's applications in scenarios such as in-context learning and Retrieval-Augmented Generation (RAG). Recent studies attempt to obtain optimal or suboptimal arrangements based on statistical results or using dataset-based search, but these methods increase inference overhead while leaving the model's inherent order bias unresolved. Other studies mitigate order sensitivity through supervised fine-tuning using augmented training sets with multiple order variants, but often at the cost of accuracy, trapping the model in consistent yet incorrect hallucinations. In this paper, we propose \textbf{D}ual \textbf{G}roup \textbf{A}dvantage \textbf{O}ptimization (\textbf{DGAO}), which aims to improve model accuracy and order stability simultaneously. DGAO calculates and balances intra-group relative accuracy advantage and inter-group relative stability advantage, rewarding the policy model for generating order-stable and correct outputs while penalizing order-sensitive or incorrect responses. This marks the first time reinforcement learning has been used to mitigate LLMs' order sensitivity. We also propose two new metrics, Consistency Rate and Overconfidence Rate, to reveal the pseudo-stability of previous methods and guide more comprehensive evaluation. Extensive experiments demonstrate that DGAO achieves superior order fairness while improving performance on RAG, mathematical reasoning, and classification tasks. Our code is available at: \url{https://github.com/Hyalinesky/DGAO}.
\end{abstract}

\section{Introduction}

Large Language Models (LLMs) achieve remarkable capabilities in utilizing input text to solve tasks~\cite{hurst2024gpt,guo2025deepseek,yang2025qwen3}, playing crucial roles in recommendation systems~\cite{lin2024data,du2024enhancing}, search systems~\cite{narayanan2025search,li2025towards}, and domain-specific question answering~\cite{goyal2024healai,chu2025domaino1s}. In particular, LLMs demonstrate powerful capabilities in in-context learning~\cite{bertsch2024context,agarwal2024many} and Retrieval-Augmented Generation (RAG)~\cite{yu2024rankrag,he2024g} tasks, being able to extract important information from contexts composed of numerous input contents and derive answers.


\begin{figure}[t]
\centering
\includegraphics[width=0.48\textwidth]{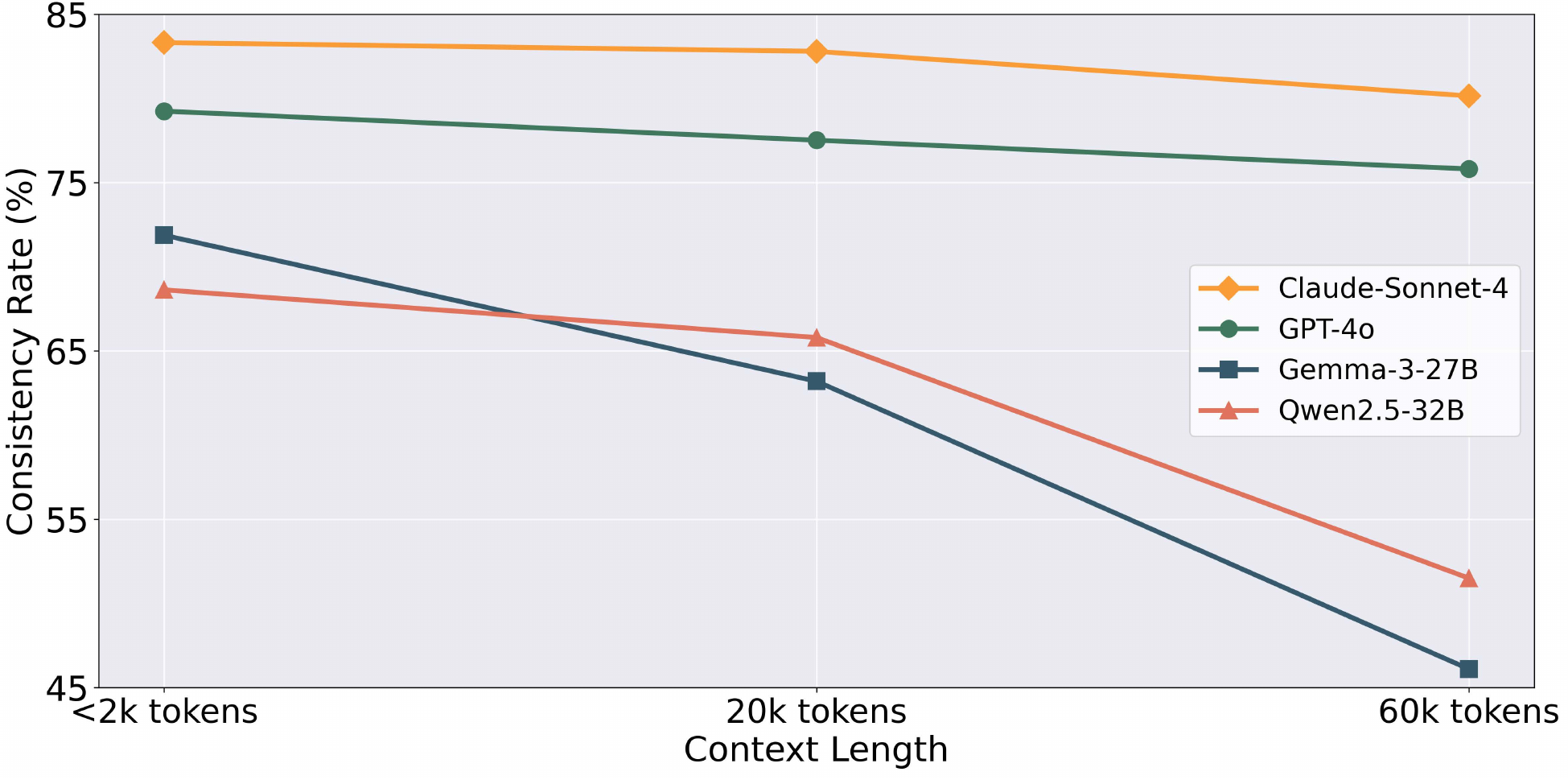} 
\caption{Consistency Rates (obtained from 8 random order variants of a sample, with higher values indicating stronger order stability) of LLMs on RAG dataset SearchQA~\cite{Dunn2017SearchQAAN} are consistently below the ideal 100\%. Moreover, as the number of tokens in the context window increases, the Consistency Rates of LLMs continue to decrease.}
\label{pic: intro}
\end{figure}

However, current mainstream LLMs face order bias~\cite{Lu2021FantasticallyOP,Liu2023LostIT,Wang2024EliminatingPB}, even for cutting-edge large-parameter LLMs, as shown in Figure~\ref{pic: intro}. This unfairness may affect their applications. In in-context learning and RAG scenarios, the prompts or documents that need to be input to LLMs can be viewed as elements in an unordered set, where the order of these elements is typically interchangeable and is not intended to influence the results~\cite{Zhang2024PositionAwarePE}. However, LLMs organize the elements in the set into natural language sequences when processing input, thereby introducing ordering to originally unordered elements. Recent research shows that LLMs are sensitive to input order. For example, statistical results indicate that many LLMs exhibit additional attention to the beginning and end of sequences~\cite{Liu2023LostIT}, and LLMs' performance typically deteriorates when key information is located in the middle of the input rather than at the beginning or end. Although placing key information in contextual positions that LLMs pay more attention to shows promise for performance improvement~\cite{tan2024order,Yao2025WhoII}, defining key information such as key frames or golden documents is very difficult in many cases. Other studies employ dataset-based search to find optimal or suboptimal arrangement orders~\cite{Lu2021FantasticallyOP,sorensen2022information,Guo2024WhatMA}, but the search process requires introducing additional inference time and typically relies on additional validation sets. These methods do not improve model structure or adjust parameters, merely utilizing the order bias phenomenon to enhance LLMs' performance on specific tasks, and cannot fundamentally mitigate the model's unfair bias.

Some studies modify the structure of LLMs to avoid or mitigate positional bias~\cite{tan2024llm,egressy2025set,chu2025graphsos}, but new structures require introducing large amounts of data for training. Orthogonally, fine-tuning models to suppress order bias shows promise.~\cite{Zhang2024PositionAwarePE} constructs multiple order variants for each training sample and uses these variants as new samples for supervised fine-tuning (SFT) of LLMs. However, as shown in Table~\ref{tab:intro}, although the model's order stability improves, its accuracy decreases. 
This reveals a disturbing phenomenon: simply fine-tuning models through data augmentation may cause models to generate consistently incorrect answers in pursuit of response consistency, which can be harmful to downstream tasks.

\begin{table}[t]
\centering
\resizebox{0.48\textwidth}{!}{
\begin{tabular}{llcc}
\toprule
\textbf{Dataset} & \textbf{Model} & \textbf{Accuracy} & \textbf{Consistency Rate} \\
\midrule
\multirow{4}{*}{SearchQA} & Llama-3.2-3B-SFT & 58.76 & 81.21 \\
& Llama-3.2-3B-PAFT & 56.31 (-2.45) & 86.48 (+5.27) \\
& Gemma-3-4B-SFT & 59.71 & 78.09 \\
& Gemma-3-4B-PAFT & 58.72 (-0.99) & 83.86 (+5.77) \\
\midrule
\multirow{4}{*}{GSM8K} & Llama-3.2-3B-SFT & 22.81 & 56.85 \\
& Llama-3.2-3B-PAFT & 14.43 (-8.38) & 58.99 (+2.14) \\
& Gemma-3-4B-SFT & 15.24 & 44.42 \\
& Gemma-3-4B-PAFT & 12.19 (-3.05) & 48.31 (+3.89) \\
\bottomrule
\end{tabular}
}
\caption{The accuracy and answer Consistency Rate of Llama-3.2-3B and Gemma-3-4B on the RAG dataset SearchQA and math dataset GSM8K constructed with 8 random orders. Position Aware Fine Tuning (PAFT) refers to augmenting each sample in the training set into 8 order variants for fine-tuning. Consistency Rate represents the maximum consistency rate of answers obtained from 8 random order variants of a sample, with higher values indicating stronger order stability. $+/-$ indicates improvement or decline relative to the SFT baseline.}
\label{tab:intro}
\end{table}

In this paper, to the best of our knowledge, this is the first time reinforcement learning is used to enhance LLMs' order stability while simultaneously improving their accuracy. We propose \textbf{D}ual \textbf{G}roup \textbf{A}dvantage \textbf{O}ptimization (\textbf{DGAO}), an on-policy reinforcement learning algorithm that rewards the policy model for generating order-stable and correct outputs and rejects order-sensitive or incorrect outputs, thereby avoiding the model accuracy decline brought by SFT with data augmentation. DGAO calculates and balances intra-group relative advantage and inter-group relative advantage, where intra-group relative advantage rewards outputs with correct answers, while inter-group relative advantage rewards order-stable groups. We also propose two new metrics: Consistency Rate, which evaluates response consistency, and Overconfidence Rate, which measures how much a model's excessive focus on response consistency affects its accuracy. Extensive experiments on three different model architectures of LLMs across mathematics, sentiment classification, and RAG tasks demonstrate that DGAO enhances model accuracy and reduces order sensitivity. Our contributions can be summarized as:


• We reveal a key pitfall in existing mitigation strategies: merely improving order stability may harm accuracy. Current SFT-based training strategies lack simultaneous, explicit supervision for the two objectives, making it difficult to optimize stability and accuracy concurrently.

• We propose DGAO, a reinforcement learning algorithm that rewards LLMs for generating order-stable and correct outputs. This is the first time reinforcement learning is used to enhance LLM performance and order stability simultaneously.

• We propose Consistency Rate and Overconfidence Rate for evaluating LLMs' order stability. Experiments show that LLMs reinforced with DGAO achieve simultaneous improvements in accuracy and order stability across mathematics, sentiment classification, and RAG tasks, with LLMs gradually moving toward order fairness.

\section{Related Works}
\subsection{Order Sensitivity in LLMs}
Current mainstream LLMs face the problem of order sensitivity~\cite{Lu2021FantasticallyOP,Liu2023LostIT,Wang2024EliminatingPB}, where when the order of elements in the input context changes, even though it is not intended to affect the results, model outputs vary in quality. Previous mainstream improvement strategies include prompting, inference-time intervention, dataset-based search, model architecture adjustment, and supervised fine-tuning. Methods based on prompting~\cite{zhang2024can,jiao2024enhancing,li2024split} or inference-time intervention~\cite{Wang2024EliminatingPB,yu2024mitigate,zhou2024unibias,wei2024unveiling} lack data-driven training. \cite{zhang2024can} proves that LLMs lack relative position awareness capability, and prompting methods that rely on contextual position perception such as~\cite{jiao2024enhancing,li2024split} may fail. Inference-time intervention methods \cite{Wang2024EliminatingPB,wei2024unveiling} often rely on heuristic assumptions and lack generalizability, potentially introducing biases due to factors such as attention score interference and hallucination.

Dataset-based search~\cite{Lu2021FantasticallyOP,sorensen2022information,Guo2024WhatMA} aims to find an optimal or sub-optimal example arrangement order to improve model performance. But they require additional forward inference~\cite{Guo2024WhatMA} and creating or introducing domain datasets~\cite{Lu2021FantasticallyOP,sorensen2022information}, thereby increasing overhead.

Model architecture adjustments suppress order sensitivity by adding additional components~\cite{tan2024llm,chu2025graphsos} or improving transformer structures~\cite{egressy2025set}. \cite{egressy2025set} removes sequential positional encoding and causal masks from transformers, adding set positional encoding and set attention masks to make the entire model's output invariant to option order permutation. \cite{chu2025graphsos} designs and trains an order selection module that organizes text-attributed graphs into better order for LLMs to better handle graph tasks. However, new model architectures require additional training, which typically involves greater training overhead than fine-tuning, thus introducing substantial training costs.

Supervised fine-tuning Methods, such as~\cite{Zhang2024PositionAwarePE}, do achieve improvements on certain tasks. It construct each training sample into multiple order variants and use these variants as new data for simultaneous training. However, as shown in Table~\ref{tab:intro}, this approach harms answer accuracy, which does not align with the ideal of guiding models to answer in an order-fair manner.

\subsection{Reinforcement Learning in LLM}
Introducing reinforcement learning to enhance LLM capabilities has achieved superior results in domains such as mathematics~\cite{Luong2024ReFTRW,shao2024deepseekmath,yang2024qwen2,ying2024internlm,chen2024step} and code~\cite{hui2024qwen2,jiao2024preference,Li2024IRCoCoIR,chen2025acereason}. Reinforcement learning for LLM fine-tuning can be categorized into On-Policy~\cite{schulman2017proximal,ouyang2022training,shao2024deepseekmath,ding2025arm} and Off-Policy~\cite{rafailov2023direct,amini2024direct,chen2024step,yuan2024self} based on whether the learning samples are generated by the current model being optimized.

Off-Policy algorithms, such as DPO~\cite{rafailov2023direct}, rely on a static, pre-collected preference dataset. The model uses a fixed dataset to learn throughout the entire training process and does not need to generate new samples during training. Off-Policy algorithms have high sample efficiency but are prone to distribution shift and are limited by preference dataset quality.

On-Policy algorithms generate optimization data in real-time from the current policy model, enabling iterative optimization of the policy model. PPO~\cite{schulman2017proximal,ouyang2022training} trains reward models and value models to perform advantage estimation for each generation and update parameters. GRPO~\cite{shao2024deepseekmath} abandons training value models and instead calculates intra-group relative advantages to guide parameter updates. However, these algorithms cannot handle one situation: how to optimize when an entire group deserves to be rewarded or rejected. For example, \textit{when generations within a group are all order-stable, how to design rewards and advantages to guide the model to learn this order-stable generation?}

\section{Methodology}
\begin{figure}[t]
\centering
\includegraphics[width=0.48\textwidth]{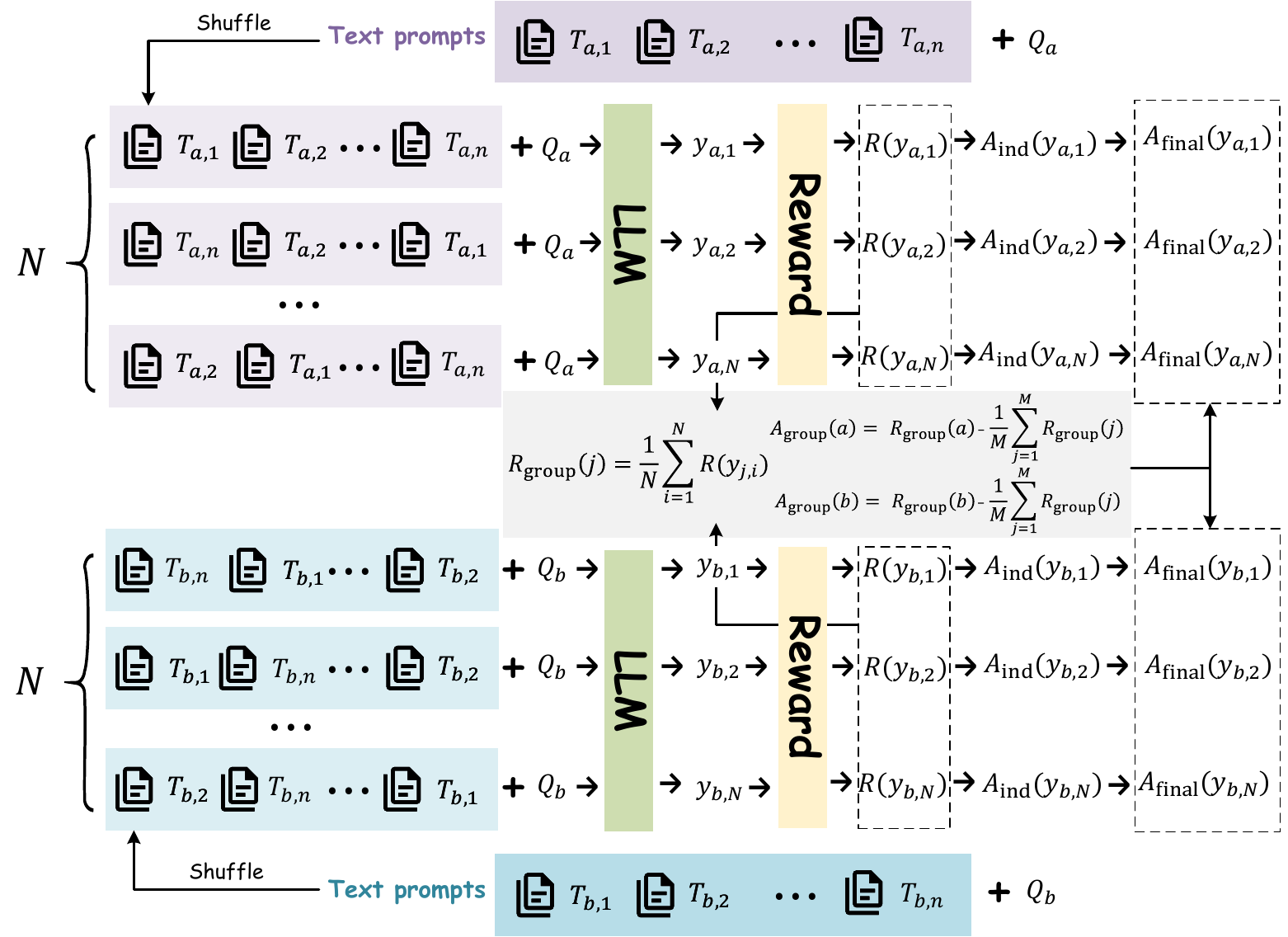} 
\caption{DGAO algorithm framework. A batch contains $M$ (e.g., $M=2$ in the figure) queries, each query consists of text prompts and $Q$. Each text prompt is constructed into $N$ order variants, and rewards are obtained through respective inference. Based on the rewards, intra-group relative advantage $A_{\text{ind}}$ and inter-group relative advantage $A_{\text{group}}$ are calculated respectively, and the aggregated final advantage $A_\text{final}$ is obtained to guide parameter updates.}
\label{pic: dgao}
\end{figure}

In this section, we introduce Dual Group Advantage Optimization (DGAO). The algorithmic framework is shown in Figure~\ref{pic: dgao}. DGAO simultaneously calculates intra-group relative accuracy advantages and inter-group relative stability advantages, thereby encouraging model to generate order-stable and correct outputs.

\subsection{Preliminary}
Reinforcement learning that calculates intra-group advantages has achieved success in recent work~\cite{guo2025deepseek,chu2025qwen}, which removes value models, thereby reducing training overhead and optimization difficulty. 
For example, Group Relative Policy Optimization (GRPO)~\cite{shao2024deepseekmath} used to train DeepSeek-R1-Zero~\cite{guo2025deepseek}, based on a simple yet effective rule: for the same problem, generate several candidate answers, called a group, and directly compare the relative merits of candidate answers within the group. Specifically, for a given query $q$, GRPO first uses the current policy $\pi_{\text{old}}$ to generate $G$ different responses ${y_1, y_2, \dots, y_G}$ and obtains corresponding rewards ${r_1, r_2, \dots, r_G}$. GRPO calculates the relative advantage of each answer through intra-group normalization:

{\small
\begin{equation}
A_i = \frac{r_i - \text{mean}(\{r_1, \dots, r_G\})}{\text{std}(\{r_1, \dots, r_G\})},
\end{equation}
}
where $A_i$ represents the advantage degree of the $i$-th answer relative to other answers within the group. Based on these relative advantage values, GRPO's objective function is:

{\small
\begin{multline}
\mathcal{L}_{GRPO}(\theta) = \frac{1}{G} \sum_{i=1}^{G} \bigg(\min \bigg( \frac{\pi_{\theta}(y_i|q)}{\pi_{\text{old}}(y_i|q)} A_i, \\
\text{clip}\left(\frac{\pi_{\theta}(y_i|q)}{\pi_{\text{old}}(y_i|q)}, 1-\epsilon, 1+\epsilon\right) A_i \bigg)- \beta\mathbb{D}_{KL} (\pi_{\theta}||\pi_{ref})\bigg),
\end{multline}
}

{\small
\begin{equation}
\mathbb{D}_{KL} (\pi_{\theta}||\pi_{ref}) = \frac{\pi_{ref}(y_i|q)}{\pi_{\theta}(y_i|q)} - \log \frac{\pi_{ref}(y_i|q)}{\pi_{\theta}(y_i|q)} - 1,
\end{equation}
}
where $\epsilon$ and $\beta$ respectively control the clipping range and the strength of KL divergence penalty. $\pi_{ref}$ is the reference policy, used to prevent the optimized policy $\pi_{\theta}$ from deviating too far and causing catastrophic forgetting.

\subsection{Order Sensitivity Phenomenon}

In in-context learning and RAG scenarios, the text prompts that serve as LLM inputs are typically unordered and are not intended to influence results through order changes. For example, documents retrieved by RAG systems (unsorted), or prompt examples in in-context learning. These text prompts can be viewed as elements in an unordered set. Given a query consisting of a set $S=\{T_1;T_2;\dots;T_n\}$ composed of contextual text prompts and a question $Q$, let $\pi$ be a permutation of the indices $\{1, \ldots, n\}$. The input sequence in a specific order is represented as $q_\pi = T_{\pi(1)} \oplus T_{\pi(2)} \oplus \cdots \oplus T_{\pi(n)} \oplus Q.$

\begin{figure}[t]
\centering
\includegraphics[width=0.48\textwidth]{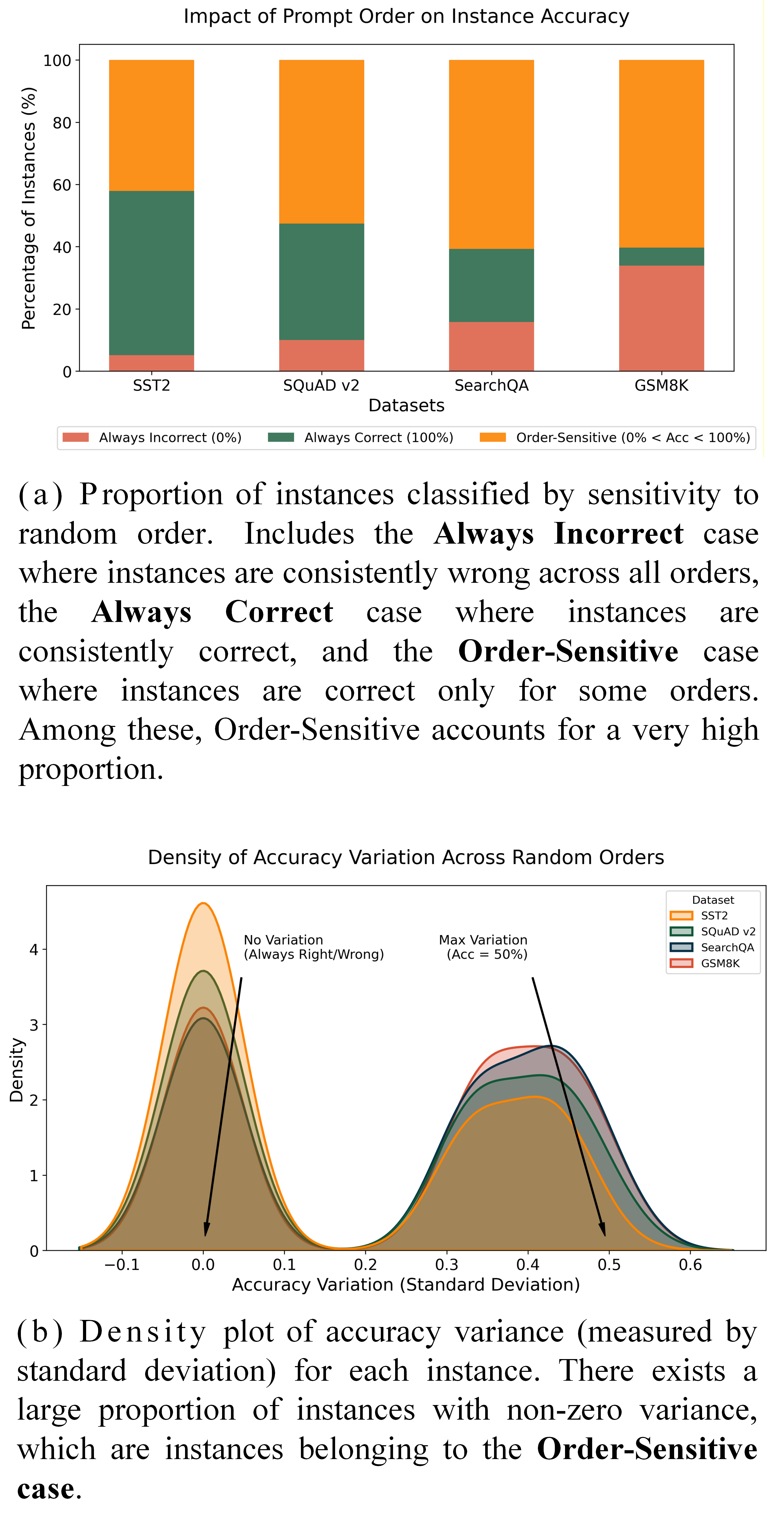} 
\caption{Impact of input order on the performance of Llama-3.2-3B across 8 random orders.}
\label{pic: orderbias}
\end{figure}

Since the elements in set $S$ are inherently unordered, ideally, a robust model should exhibit permutation invariance, i.e., for any permutations $\pi, \pi'$, $P(y|q_\pi) = P(y|q_{\pi'}).$
However, LLMs model the joint probability of the output sequence $y = (y_1, \ldots, y_L)$ through the chain rule:

{\small
\begin{equation}
P(y|q_\pi) = \prod_{t=1}^{L} P(y_t | y_{<t}, q_\pi).
\end{equation}
}
Due to the positional embeddings and causal attention mechanisms in mainstream Transformer-based LLMs, the hidden state representation $h_t$ at step $t$ depends on the specific absolute and relative positions of tokens in $q_\pi$. Therefore, changes in the context order $\pi$ alter the attention weights and embedding representations by changing token positions, which may lead to differences in conditional probability distributions, i.e., $P(y|q_\pi) \neq P(y|q_{\pi'}).$
This difference manifests as the model assigning different probabilities to correct answers under different orders, thereby producing the \textit{\textbf{order bias}} phenomenon, as shown in Figure~\ref{pic: orderbias}. This phenomenon can lead to unfairness and affect downstream tasks. Moreover, recent research indicates that there is no unified optimal order across models or tasks~\cite{Lu2021FantasticallyOP}, and even the "Lost in the middle" phenomenon, where models focus on the beginning and end of contexts does not always occur~\cite{Zhang2024PositionAwarePE,tan2024order}. This brings challenges to the application of methods that directly order based on statistical preferences.

\subsection{Dual Group Advantage Optimization}
To mitigate order sensitivity, we aim to optimize the policy $\pi_\theta$ so that it consistently generates correct answers under different permutations. We theoretically analyze the limitations of existing strategies in Appendix \ref{Appendix: Theoretical Motivation}, including standard SFT on augmented data (PAFT~\cite{Zhang2024PositionAwarePE}) and RL approaches, in addressing this problem.


To better address order sensitivity, we propose \textbf{D}ual \textbf{G}roup \textbf{A}dvantage \textbf{O}ptimization (\textbf{DGAO}), with the algorithm architecture shown in Figure~\ref{pic: dgao}. DGAO is an On-Policy reinforcement learning algorithm that aims to reward the policy model for generating order-stable and correct outputs while rejecting order-sensitive or incorrect outputs, thereby avoiding the model accuracy degradation caused by data augmentation SFT. Unlike GRPO, which only considers intra-group relative advantage, DGAO calculates and balances both intra-group relative advantage and inter-group relative advantage, where intra-group relative advantage rewards outputs with correct answers, while inter-group relative advantage rewards order-stable groups.

The core idea of DGAO is based on a key construction: in a training batch, there are $M$ different queries, each query consists of $n$ text prompts and one question $Q$, i.e., $q=T_{\pi(1)} \oplus T_{\pi(2)} \oplus \cdots \oplus T_{\pi(n)} \oplus Q$. For each query $q_j$, $N$ versions with different orders of text prompts are constructed, denoted as ${q_{j,1}, q_{j,2}, \ldots, q_{j,N}}$. For each $q_{j,i}$, the model generates a result $y_{j,i}$. In this way, $N$ ordered samples and their results form a group. DGAO defines two advantage components to capture performance at different levels:

\textbf{Inter-group advantage} $A_{\text{group}}(j)$ measures the overall stability of a group by calculating the group reward $R_{\text{group}}(j) = \frac{1}{N} \sum_{i=1}^{N} R(y_{j,i})$ and global baseline $b = \frac{1}{M} \sum_{j=1}^{M} R_{\text{group}}(j)$, obtaining $A_{\text{group}}(j) = R_{\text{group}}(j) - b$. This advantage remains the same for all samples within group $j$, reflecting the performance of that group as a whole relative to the batch average. Here, the reward for individual model results $R(y_{j,i})$ can be given by a reward model or rule-based reward function. In our implementation, we use rule-based reward, where the result gets a score of 1 if correct and 0 if incorrect. It should be noted that inter-group advantage does not encourage the model to generate consistently incorrect answers within a group. Although this may appear to be "order stable", it would harm model accuracy. Therefore, groups that tend to output consistently incorrect answers also have lower inter-group advantage.

\textbf{Intra-group advantage} $A_{\text{ind}}(y_{j,i})$ inherits the relative advantage idea from GRPO and measures the accuracy performance of individual generated results relative to their same-group samples. By calculating $A_{\text{ind}}(y_{j,i}) = R(y_{j,i}) - R_{\text{group}}(j)$, this advantage reflects the difference of each generated result relative to the average level of its group, may have different values for different samples within group $j$. 

DGAO uses hyperparameter $\alpha \in [0,1]$ to perform weighted combination of two advantage components, forming hybrid advantage: $A_\text{hybrid}(y_{j,i})=\alpha \cdot A_\text{group}(j)+(1-\alpha) \cdot A_\text{ind}(y_{j,i}).$
When $\alpha \to 1$, the model focuses more on order-stability performance; when $\alpha \to 0$, the model focuses more on intra-group accuracy performance. To ensure training stability, DGAO normalizes $A_{\text{hybrid}}$ for the entire batch:

{\small
\begin{equation}
    A_\text{final}(y_{j,i})=\frac{A_\text{hybrid}(y_{j,i})-\mu_A}{\sigma_A+\epsilon_A},
\end{equation}
}
where $\mu_A$ and $\sigma_A$ are the mean and standard deviation of hybrid advantage in the current batch respectively, and $\epsilon_A$ is a small constant. Finally, the objective function of DGAO can be expressed as:

{\scriptsize
\begin{multline}
\mathcal{L}_{\text{DGAO}}(\theta) = \frac{1}{MN} \sum_{j=1}^{M} \sum_{i=1}^{N} (\min ( \frac{\pi_{\theta}(y_{j,i}|q_{j,i})}{\pi_{\text{old}}(y_{j,i}|q_{j,i})} A_{\text{final}}(y_{j,i}),  \\
\text{clip}(\frac{\pi_{\theta}(y_{j,i}|q_{j,i})}{\pi_{\text{old}}(y_{j,i}|q_{j,i})}, 1-\epsilon, 1+\epsilon) A_{\text{final}}(y_{j,i}) )
- \beta\mathbb{D}_{KL} (\pi_{\theta}||\pi_{ref})).
\end{multline}
}

\section{Experiments}
In this section, we evaluate the performance of LLMs trained with DGAO in terms of accuracy and order sensitivity. We aim to address the following questions: \textbf{RQ}1: How do models perform after training with DGAO? \textbf{RQ2}: How does DGAO compare with other methods for suppressing order sensitivity? \textbf{RQ3}: How does DGAO compare with other reinforcement learning methods? \textbf{RQ4}: How do inter-group advantages and intra-group advantages affect the effectiveness of DGAO? \textbf{RQ5}: What is the generalization capability of order stability for models trained with DGAO?

\subsection{Experimental Setup}
\textbf{Datasets and metrics}. We select 3 order-sensitive tasks, mathematics, sentiment classification, and RAG, and choose datasets that are widely used for evaluation. Mathematics datasets include GSM8K~\cite{Cobbe2021TrainingVT} and CM17K~\cite{qin2021neural}.
In our experiments, each dataset samples 2K training samples, where each sample's query context contains 4 randomly sampled question-answer pairs as prompts, followed by 1 question that the model needs to answer. Then, we randomly shuffle the order between prompts to obtain 8 query variants, resulting in a total of 16K training data, with 8 variants recorded as 1 group.

For sentiment classification dataset, we choose SST2~\cite{socher2013recursive}, which consists of sentences extracted from movie reviews with each sentence annotated with a unique Positive or Negative sentiment, forming a binary classification task. In our experiments, we sample 2K training samples, where each sample's query context contains 16 question-answer pairs as prompts, followed by 1 question that the model needs to answer. Then, we randomly shuffle the order between prompts to obtain 8 query variants, resulting in a total of 16K training data, with 8 variants recorded as 1 group.

For RAG tasks, we choose SQuAD v2~\cite{rajpurkar2018know} and SearchQA~\cite{Dunn2017SearchQAAN}. Each question includes the question, answer, and documents needed to obtain the answer. In our experiments, each dataset samples 2K training samples respectively, where each sample's query context contains 5 documents, followed by 1 question that the model needs to answer. Among the 5 documents, only one is relevant to answering the current question. Then, we randomly shuffle the order between prompts to obtain 8 query variants, resulting in a total of 16K training data, with 8 variants recorded as 1 group.

Test sets for all datasets are constructed in the same way as the training sets. Examples can be found in Appendix~\ref{ap:data}.

To comprehensively evaluate the model's answer accuracy and order stability, we introduce 3 metrics. In test samples, $M$ represents the number of groups, i.e., the number of original queries. $N$ represents the number of order variants within groups, $L_j$ represents the label answer for the $j$-th group, $y_{j,i}$ represents the model's output for the $i$-th order variant query in the $j$-th group, and $I(\cdot)$ represents the indicator function. When the condition in parentheses is true, its value is 1; when the condition is false, its value is 0. The metrics are formally defined as follows:

    
    

• \textbf{Accuracy}: $\frac{1}{M \times N} \sum_{j=1}^{M} \sum_{i=1}^{N} I(y_{j,i} = L_j).$ It represents the average test accuracy of all test variants.

• \textbf{Consistency Rate}: 
    
$\frac{1}{M} \sum_{j=1}^{M} \left( \frac{1}{N} \max_{p \in \mathcal{Y}_j} \sum_{i=1}^{N} I(y_{j,i} = p) \right),$ where $\mathcal{Y}_j = \{y_{j,1}, y_{j,2}, \dots, y_{j,N}\}$ is the set of output results given by the model for all $N$ versions of the $j$-th query. This value represents the maximum consistency rate of the model's answers, with larger values indicating larger order-stability.

• \textbf{Overconfidence Rate}: 
    
$\frac{1}{M} \sum_{j=1}^{M} \left( \frac{1}{N} \max_{p \in \mathcal{Y}_j, p \neq L_j} \sum_{i=1}^{N} I(y_{j,i} = p) \right),$ where $\mathcal{Y}_j = \{y_{j,1}, y_{j,2}, \dots, y_{j,N}\}.$ If for a certain group $j$, all outputs are correct (i.e., there is no output $p$ that is not equal to $L_j$), then the value of the $max$ term for that group is 0. This value represents the maximum consistent error rate of the model's answers, with larger values indicating stronger model hallucination, as they are overconfident and tend to output certain incorrect answers.

\begin{table}[t]
\resizebox{.49\textwidth}{!}{
\begin{tabular}{llccc}
\toprule
\textbf{Model} & \textbf{Dataset} & \textbf{Accuracy $\uparrow$} & \textbf{Consistency Rate $\uparrow$} & \textbf{Overconfidence Rate $\downarrow$} \\
\midrule
\multirow{15}{*}{Qwen-2.5-3B} 
& SST2-SFT & 93.48 & 97.33 & 6.91 \\
& SST2-PAFT & 92.45 (-1.03) & 99.28 (+1.95) & 7.18 (+0.27) \\
& \textbf{SST2-DGAO} & \textbf{93.53 (+0.05)} & \textbf{99.35 (+2.02)} & \textbf{6.76 (-0.15)} \\
\cmidrule{2-5}
& SQuAD v2-SFT & 84.97 & 92.91 & 11.14 \\
& SQuAD v2-PAFT & 81.59 (-3.38) & \textbf{93.56 (+0.65)} & 13.77 (+2.63) \\
& \textbf{SQuAD v2-DGAO} & \textbf{85.63 (+0.66)} & 93.35 (+0.44) & \textbf{10.22 (-0.92)} \\
\cmidrule{2-5}
& SearchQA-SFT & 68.49 & 85.54 & 20.64 \\
& SearchQA-PAFT & 69.62 (+1.13) & \textbf{90.91 (+5.37)} & 24.10 (+3.46) \\
& \textbf{SearchQA-DGAO} & \textbf{71.87 (+3.38)} & 87.81 (+2.27) & \textbf{19.26 (-1.38)} \\
\cmidrule{2-5}
& GSM8k-SFT & 42.53 & 67.50 & 28.52 \\
& GSM8k-PAFT & 33.81 (-8.72) & 67.97 (+0.47) & 36.99 (+8.47) \\
& \textbf{GSM8k-DGAO} & \textbf{42.91 (+0.38)} & \textbf{68.11 (+0.61)} & \textbf{28.41 (-0.11)} \\
\cmidrule{2-5}
& CM17k-SFT & 28.33 & 78.02 & 51.24 \\
& CM17k-PAFT & 18.83 (-9.50) & \textbf{83.20 (+5.18)} & 64.94 (+13.70) \\
& \textbf{CM17k-DGAO} & \textbf{29.35 (+1.02)} & 77.24 (-0.78) & \textbf{46.99 (-4.25)} \\
\midrule
\multirow{15}{*}{Llama-3.2-3B}
& SST2-SFT & 89.47 & 99.08 & 10.26 \\
& SST2-PAFT & 87.62 (-1.85) & 99.08 (+0.00) & 11.87 (+1.61) \\
& \textbf{SST2-DGAO} & \textbf{90.64 (+1.17)} & \textbf{99.14 (+0.06)} & \textbf{8.80 (-1.46)} \\
\cmidrule{2-5}
& SQuAD v2-SFT & 73.73 & 88.98 & 18.54 \\
& SQuAD v2-PAFT & 58.11 (-15.62) & 82.62 (-6.36) & 29.19 (+10.65) \\
& \textbf{SQuAD v2-DGAO} & \textbf{75.44 (+1.71)} & \textbf{89.60 (+0.62)} & \textbf{17.57 (-0.97)} \\
\cmidrule{2-5}
& SearchQA-SFT & 58.72 & 81.21 & 25.88 \\
& SearchQA-PAFT & 56.34 (-2.38) & \textbf{86.45 (+5.24)} & 33.49 (+7.61) \\
& \textbf{SearchQA-DGAO} & \textbf{62.51 (+3.79)} & 84.72 (+3.51) & \textbf{25.43 (-0.45)} \\
\cmidrule{2-5}
& GSM8k-SFT & 22.85 & 56.78 & 37.56 \\
& GSM8k-PAFT & 14.36 (-8.49) & 58.00 (+1.22) & 46.87 (+9.31) \\
& \textbf{GSM8k-DGAO} & \textbf{23.19 (+0.34)} & \textbf{58.47 (+1.69)} & \textbf{34.72 (-2.84)} \\
\cmidrule{2-5}
& CM17k-SFT & 15.08 & 70.45 & 56.80 \\
& CM17k-PAFT & 10.00 (-5.08) & \textbf{77.12 (+6.67)} & 67.96 (+11.16) \\
& \textbf{CM17k-DGAO} & \textbf{15.19 (+0.11)} & 70.61 (+0.16) & \textbf{51.14 (-5.66)} \\
\midrule
\multirow{15}{*}{Gemma-3-4B}
& SST2-SFT & 89.98 & 98.95 & \textbf{9.27} \\
& SST2-PAFT & 88.64 (-1.34) & \textbf{99.66 (+0.71)} & 11.25 (+1.98) \\
& \textbf{SST2-DGAO} & \textbf{90.11 (+0.13)} & 99.03 (+0.08) & 9.48 (+0.21) \\
\cmidrule{2-5}
& SQuAD v2-SFT & 74.09 & 85.31 & 16.29 \\
& SQuAD v2-PAFT & 63.92 (-10.17) & 84.58 (-0.73) & 24.80 (+8.51) \\
& \textbf{SQuAD v2-DGAO} & \textbf{75.82 (+1.73)} & \textbf{85.49 (+0.18)} & \textbf{14.80 (-1.49)} \\
\cmidrule{2-5}
& SearchQA-SFT & 59.72 & 78.09 & 23.00 \\
& SearchQA-PAFT & 58.64 (-1.08) & \textbf{83.82 (+5.73)} & 29.74 (+6.74) \\
& \textbf{SearchQA-DGAO} & \textbf{63.10 (+3.38)} & 80.60 (+2.51) & \textbf{22.11 (-0.89)} \\
\cmidrule{2-5}
& GSM8k-SFT & 15.23 & 44.39 & 31.78 \\
& GSM8k-PAFT & 12.17 (-3.06) & \textbf{48.27 (+3.88)} & 38.66 (+6.88) \\
& \textbf{GSM8k-DGAO} & \textbf{15.59 (+0.36)} & 44.43 (+0.04) & \textbf{31.43 (-0.35)} \\
\cmidrule{2-5}
& CM17k-SFT & 11.39 & 59.32 & 49.26 \\
& CM17k-PAFT & 9.54 (-1.85) & \textbf{70.18 (+10.86)} & 52.84 (+3.58) \\
& \textbf{CM17k-DGAO} & \textbf{11.78 (+0.39)} & 59.57 (+0.25) & \textbf{47.12 (-2.14)} \\
\bottomrule
\end{tabular}
}
\caption{Accuracy and order stability metrics of LLMs with different training methods. $+/-$ indicates improvement or degradation relative to the SFT baseline. $\uparrow$ indicates that higher values of this metric represent better model accuracy or stronger order stability, $\downarrow$ indicates that lower values of this metric represent lower model hallucination.}
\label{tab:sftvsdgao}
\end{table}

\textbf{Baselines}. The main baselines for our method are SFT and Position Aware Fine Tuning (PAFT)~\cite{Zhang2024PositionAwarePE}. SFT refers to standard fine-tuning using original training samples. PAFT refers to fine-tuning using datasets augmented with 8 order variants. Additionally, although orthogonal to our work, we also compare other baselines that mitigate order sensitivity. Since our method does not change the model architecture, we mainly compare the following methods:


• Prompt: Adding "please do not consider the order of text when answering" in the query.

• Lost in the middle~\cite{Liu2023LostIT}: Placing gold document at beginning or end of the context.

• PINE~\cite{wang2024eliminating}: Modify attention mechanism and reassign positional information.

• SPHS~\cite{yu2024mitigate}: Adjust the activation values of position-related hidden states.

• GlobalE~\cite{Lu2021FantasticallyOP} and DEmO~\cite{Guo2024WhatMA}: Methods for searching optimal or suboptimal orders. Although they are orthogonal to our method, we still include them for comparison.

Additionally, since we introduce a new reinforcement learning method, we also compare the effects of traditional reinforcement learning methods, including PPO~\cite{ouyang2022training} and GRPO~\cite{guo2025deepseek}.

\textbf{Implementation Details}. We evaluate 3 models: Qwen-2.5-3B~\cite{qwen2_5}, Llama-3.2-3B~\cite{llama32_docs}, Gemma-3-4B~\cite{Kamath2025Gemma3T}. For SFT and PAFT, all models use 2K and 16K data of each task respectively for full-parameter fine-tuning. For our DGAO, we first use half of the 16K augmented data for full-parameter SFT to cold-start the model, then use the other half of the data for DGAO reinforcement fine-tuning. $\alpha$ and $\beta$ are set to 0.5 and 0.05, respectively with detailed parameter settings in Appendix~\ref{ap:exp}.


\subsection{Main Results}
\label{mainresults}
Table~\ref{tab:sftvsdgao} shows the accuracy, Consistency Rate, and Overconfidence Rate metrics of LLMs trained with SFT, PAFT, and DGAO. Compared to SFT, although only variants with different orders are used and no additional data is introduced, DGAO-trained models improve accuracy and Consistency Rate on almost all tasks while reducing Overconfidence Rate. While PAFT using the same training data improves Consistency Rate across nearly all tasks, it reduces accuracy and increases Overconfidence Rate. This indicates that PAFT introduces more hallucinations to maintain answer consistency, thereby harming the responses. Although DGAO does not always lead PAFT in Consistency Rate, it performs better in both accuracy and Overconfidence Rate, making it more reliable in practical applications. Additionally, experimental results on the scalability of DGAO on larger parameter LLMs can be found in Appendix~\ref{app:larger scale}.

\begin{table*}[t]
\resizebox{\textwidth}{!}{%
\begin{tabular}{l|ccc|ccc|ccc|ccc|ccc}
\toprule
\multicolumn{1}{c}{\multirow{2}{*}{\textbf{Method}}} & \multicolumn{3}{c}{\textbf{SST2}}                                                                                      & \multicolumn{3}{c}{\textbf{SQuAD}}                                                                                     & \multicolumn{3}{c}{\textbf{SearchQA}}                                                                                  & \multicolumn{3}{c}{\textbf{GSM8K}}                                       & \multicolumn{3}{c}{\textbf{CM17K}}                                       \\
\cmidrule{2-16}
\multicolumn{1}{c}{}                        & \multicolumn{1}{c}{Accuracy $\uparrow$} & \multicolumn{1}{c}{CR $\uparrow$} & \multicolumn{1}{c}{OR $\downarrow$} & \multicolumn{1}{c}{Accuracy $\uparrow$} & \multicolumn{1}{c}{CR $\uparrow$} & \multicolumn{1}{c}{OR $\downarrow$} & \multicolumn{1}{c}{Accuracy $\uparrow$} & \multicolumn{1}{c}{CR $\uparrow$} & \multicolumn{1}{c}{OR $\downarrow$} & \multicolumn{1}{c}{Accuracy $\uparrow$} & \multicolumn{1}{c}{CR $\uparrow$} & \multicolumn{1}{c}{OR $\downarrow$}    & \multicolumn{1}{c}{Accuracy $\uparrow$} & \multicolumn{1}{c}{CR $\uparrow$} & \multicolumn{1}{c}{OR $\downarrow$}    \\
\midrule
SFT                                         & 89.47                        & 99.08                                & 10.26                                   & 73.73                        & 88.98                                & 18.54                                   & 58.72                        & 81.21                                & 25.88                                   & 22.85                  & 56.78                  & 37.56                  & 15.08                  & 70.45                  & 56.80                  \\
Prompt                                      & 89.28 (-0.19)                & 99.03 (-0.05)                        & 10.52 (+0.26)                           & 73.37 (-0.36)                & 89.14 (+0.16)                        & 18.90 (+0.36)                           & 58.35 (-0.37)                & 81.08 (-0.13)                        & 26.21 (+0.33)                           & 22.99 (+0.14)          & 56.59 (-0.19)          & 37.81 (+0.25)          & 14.93 (-0.15)          & 70.68 (+0.23)          & 57.03 (+0.23)          \\
Pos-Beginning                               & -                & -                       & -                          & 75.21 (+1.48)                & \textbf{92.35 (+3.37)}                        & \textbf{15.85 (-2.69)}                           & 62.49 (+3.77)                & \textbf{87.99 (+6.78)}                        & 26.24 (+0.36)                           & -         & -          & -          & -          & -          & -          \\
Pos-End                                     & -                & -                       & -                            & 74.99 (+1.26)                & 90.49 (+1.51)                        & 17.65 (-0.89)                           & 62.24 (+3.52)                & 87.41 (+6.20)                        & 26.52 (+0.64)                           & -          & -          & -          & -          & -          & -          \\
PINE                                        & 84.15 (-5.32)                & 91.51 (-7.57)                        & 19.86 (+9.60)                           & 66.91 (-6.82)                & 81.75 (-7.23)                        & 18.42 (-0.12)                           & 52.23 (-6.49)                & 75.31 (-5.90)                        & 26.09 (+0.21)                           & 15.49 (-7.36)          & 49.88 (-6.90)          & 38.19 (+0.63)          & 7.92 (-7.16)           & 62.25 (-8.20)          & 56.88 (+0.08)          \\
SPHS                                        & 58.85 (-30.62)               & 72.27 (-26.81)                       & 23.85 (+13.59)                          & 73.59 (-0.14)                & 89.10 (+0.12)                        & 18.71 (+0.17)                           & 47.78 (-10.94)               & 67.65 (-13.56)                       & 26.22 (+0.34)                           & 22.48 (-0.37)          & 56.12 (-0.66)          & 37.38 (-0.18)          & 14.99 (-0.09)          & 70.36 (-0.09)          & 57.17 (+0.37)          \\
GlobalE                                     & 89.31 (-0.16)                & 99.01 (-0.07)                        & 10.49 (+0.23)                           & 73.81 (+0.08)                & 89.15 (+0.17)                        & 18.30 (-0.24)                           & 58.88 (+0.16)                & 81.19 (-0.02)                        & 25.79 (-0.09)                           & 22.89 (+0.04)          & 56.85 (+0.07)          & 37.51 (-0.05)          & 14.88 (-0.20)          & 70.49 (+0.04)          & 57.06 (+0.26)          \\
DEmO                                        & 89.52 (+0.05)                & 99.11 (+0.03)               & 10.12 (-0.14)                           & 74.34 (+0.61)                & 89.25 (+0.27)                        & 18.25 (-0.29)                           & 59.26 (+0.54)                & 81.93 (+0.72)                        & 25.58 (-0.30)                           & 22.95 (+0.10)          & 56.91 (+0.13)          & 37.45 (-0.11)          & 15.15 (+0.07)          & 70.51 (+0.06)          & 56.71 (-0.09)          \\
\textbf{DGAO}                               & \textbf{90.64 (+1.17)}       & \textbf{99.14 (+0.06)}                        & \textbf{8.80 (-1.46)}                   & \textbf{75.44 (+1.71)}       & 89.60 (+0.62)                        & 17.57 (-0.97)                           & \textbf{62.51 (+3.79)}       & 84.72 (+3.51)                        & \textbf{25.43 (-0.45)}                  & \textbf{23.19 (+0.34)} & \textbf{58.47 (+1.69)} & \textbf{34.72 (-2.84)} & \textbf{15.19 (+0.11)} & \textbf{70.61 (+0.16)} & \textbf{51.14 (-5.66)} \\
\bottomrule
\end{tabular}%
}
\caption{Performance comparison of different methods on Llama-3.2-3B. CR represents Consistency Rate, OR represents Overconfidence Rate.}
\label{tad: res_methods}
\end{table*}

\begin{table*}[t]
\centering
\resizebox{1.0\textwidth}{!}{%
\begin{tabular}{l|ccc|ccc|ccc|ccc|ccc}
\toprule
\multicolumn{1}{c}{\multirow{2}{*}{\textbf{Method}}} & \multicolumn{3}{c}{\textbf{SST2}}                                                                                      & \multicolumn{3}{c}{\textbf{SQuAD}}                                                                                     & \multicolumn{3}{c}{\textbf{SearchQA}}                                                                                  & \multicolumn{3}{c}{\textbf{GSM8K}}                                       & \multicolumn{3}{c}{\textbf{CM17K}}                                       \\
\cmidrule{2-16}
\multicolumn{1}{c}{}                        & \multicolumn{1}{c}{Accuracy $\uparrow$} & \multicolumn{1}{c}{CR $\uparrow$} & \multicolumn{1}{c}{OR $\downarrow$} & \multicolumn{1}{c}{Accuracy $\uparrow$} & \multicolumn{1}{c}{CR $\uparrow$} & \multicolumn{1}{c}{OR $\downarrow$} & \multicolumn{1}{c}{Accuracy $\uparrow$} & \multicolumn{1}{c}{CR $\uparrow$} & \multicolumn{1}{c}{OR $\downarrow$} & \multicolumn{1}{c}{Accuracy $\uparrow$} & \multicolumn{1}{c}{CR $\uparrow$} & \multicolumn{1}{c}{OR $\downarrow$}    & \multicolumn{1}{c}{Accuracy $\uparrow$} & \multicolumn{1}{c}{CR $\uparrow$} & \multicolumn{1}{c}{OR $\downarrow$}    \\
\midrule
SFT & 89.47 & 99.08 & 10.26 & 73.73 & 88.98 & 18.54 & 58.72 & 81.21 & 25.88 & 22.85 & 56.78 & 37.56 & 15.08 & 70.45 & 56.80 \\
w/o $A_{\text{group}}$ & 90.29 (+0.82) & 98.83 (-0.25) & 9.39 (-0.87) & 75.41 (+1.68) & 89.16 (+0.18) & \textbf{17.45 (-1.09)} & 61.98 (+3.26) & 80.30 (-0.91) & 25.51 (-0.37) & 23.09 (+0.24) & 56.27 (-0.51) & 37.64 (+0.08) & \textbf{15.25 (+0.17)} & 70.27 (-0.18) & 56.31 (-0.49) \\
w/o $A_{\text{ind}}$ & 89.44 (-0.03) & 99.04 (-0.04) & 10.08 (-0.18) & 73.92 (+0.19) & 88.56 (-0.42) & 18.32 (-0.22) & 57.82 (-0.90) & 80.39 (-0.82) & 26.19 (+0.31) & 21.42 (-1.43) & 55.50 (-1.28) & 37.43 (-0.13) & 14.27 (-0.81) & 62.19 (-8.26) & \textbf{49.74 (-7.06)} \\
\textbf{DGAO} & \textbf{90.64 (+1.17)} & \textbf{99.14 (+0.06)} & \textbf{8.80 (-1.46)} & \textbf{75.44 (+1.71)} & \textbf{89.60 (+0.62)} & 17.57 (-0.97) & \textbf{62.51 (+3.79)} & \textbf{84.72 (+3.51)} & \textbf{25.43 (-0.45)} & \textbf{23.19 (+0.34)} & \textbf{58.47 (+1.69)} & \textbf{34.72 (-2.84)} & 15.19 (+0.11) & \textbf{70.61 (+0.16)} & 51.14 (-5.66) \\
\bottomrule
\end{tabular}%
}
\caption{Ablation study results of DGAO (using Llama-3.2-3B as the base model).}
\label{tab:ablation_compare}
\end{table*}

\begin{table}[t]
\centering
\resizebox{0.49\textwidth}{!}{%
\begin{tabular}{lcccccc}
\toprule
\multicolumn{1}{c}{\multirow{2}{*}{\textbf{Method}}} & \multicolumn{3}{c}{\textbf{SQuAD}}                                       & \multicolumn{3}{c}{\textbf{GSM8K}}                                       \\
\cmidrule(lr){2-4} \cmidrule(lr){5-7}
\multicolumn{1}{c}{}                                 & Accuracy $\uparrow$              & CR $\uparrow$                   & OR $\downarrow$                    & Accuracy $\uparrow$            & CR $\uparrow$                    & OR $\downarrow$                    \\
\midrule
SFT                                                  & 73.73                  & 88.98                  & 18.54                  & 22.85                  & 56.78                  & 37.56                  \\
PPO                                                  & 73.02 (-0.71)          & 86.90 (-2.08)          & 22.28 (+3.74)          & 22.99 (+0.14)          & 56.21 (-0.57)          & 37.66 (+0.10)          \\
GRPO                                                 & 73.11 (-0.62)          & 88.15 (-0.83)          & 22.49 (+3.95)          & 22.92 (+0.07)          & 55.70 (-1.08)          & 37.59 (+0.03)          \\
\textbf{DGAO}                                        & \textbf{75.44 (+1.71)} & \textbf{89.60 (+0.62)} & \textbf{17.57 (-0.97)} & \textbf{23.19 (+0.34)} & \textbf{58.47 (+1.69)} & \textbf{34.72 (-2.84)} \\
\bottomrule
\end{tabular}%
}
\caption{Performance comparison of different reinforcement learning methods on Llama-3.2-3B.}
\label{tab:rl_comparison}
\end{table}

Table~\ref{tad: res_methods} compares the evaluation results of other methods with those of DGAO-trained models. The accuracy and order stability metrics of Prompt are almost on par with SFT, indicating that naively prompting the model not to focus on context order cannot solve order bias. Pos-Beginning and Pos-End are two variants of "Lost in middle," which place the most relevant documents in RAG tasks at the beginning or end of the context, respectively. They serve as strong baselines because they are equivalent to transforming the test set into a form more suitable for the model to answer. Therefore, they achieve better accuracy and order stability in RAG tasks, indicating that Llama-3.2-3B indeed pays more attention to the beginning and end of the context. However, in RAG systems that lack ranking or in-context learning tasks (such as SST2, GSM8K, CM17K), it is difficult to define which prompt is the key prompt, making these methods inapplicable. PINE and SPHS lag behind DGAO by 8.05\% and 9.87\% in accuracy, respectively, and by 8.37\% and 9.42\% in Consistency Rate. It is worth noting that DGAO is actually orthogonal to the other methods in Table~\ref{tad: res_methods}, as it neither changes the model structure nor improves the decoding strategy. Therefore, these methods can be simply implemented on DGAO-trained models, potentially bringing further improvements in accuracy and order fairness. We leave this part of the research for future work.

Table~\ref{tab:rl_comparison} shows results of models trained with different reinforcement learning methods. Both PPO and GRPO reduce the Consistency Rate, indicates that traditional reinforcement learning methods cannot enhance the model's order stability. DGAO not only improves order stability but also slightly outperforms PPO and GRPO in accuracy improvement, leading to simultaneous improvements in both model accuracy and order fairness.

\subsection{Ablation Experiments}
Table~\ref{tab:ablation_compare} shows the performance of models trained with DGAO without inter-group relative advantage $A_\text{group}$ ($\alpha=0$) or without intra-group relative advantage $A_\text{ind}$ ($\alpha=1$). The results indicate that without $A_\text{group}$, there are general improvements in accuracy and decreases in Overconfidence Rate, showing that the model indeed rewards more accurate answers through intra-group relative advantage, but fails to improve Consistency Rate, thus not alleviating order sensitivity. Without $A_\text{ind}$, the model performs poorly on almost all metrics, indicating that merely rewarding inter-group relative advantage cannot effectively help the model learn. 

\begin{figure}[t]
\centering
\includegraphics[width=0.48\textwidth]{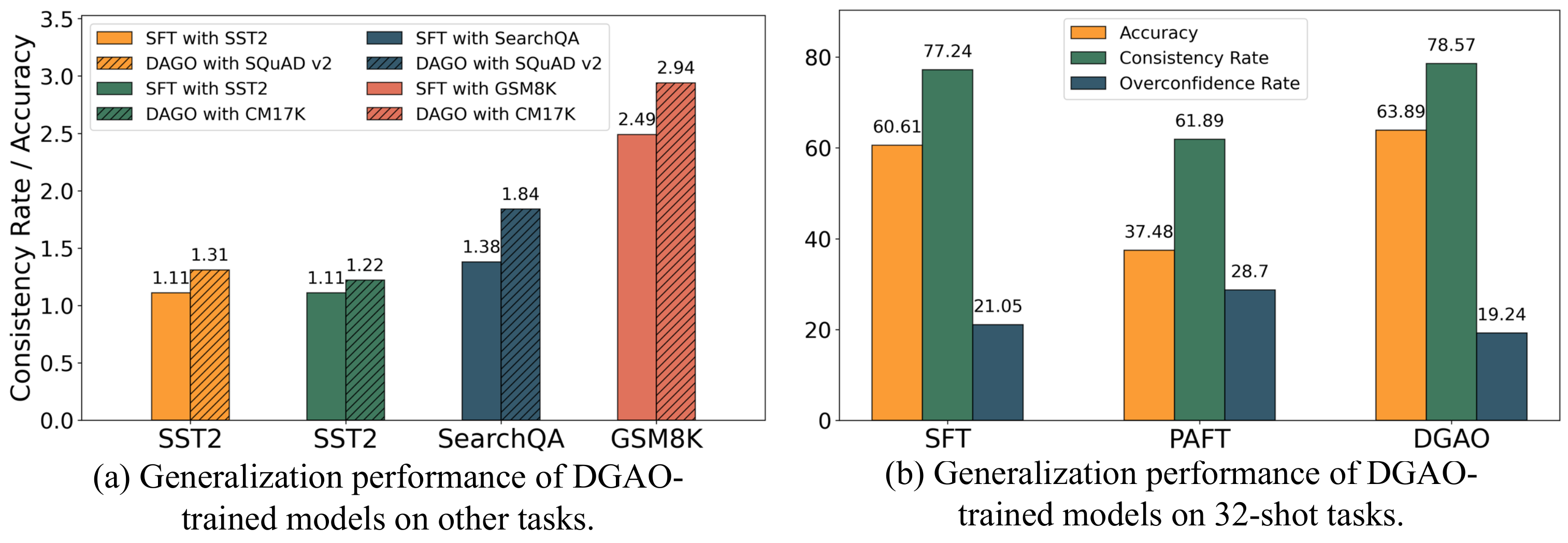} 
\caption{DGAO's Order Stability Generalization (using Llama-3.2-3B as base model).}
\label{pic: gene}
\end{figure}

\subsection{Generalization Capability}
We train DGAO on the SQuAD v2 and CM17K datasets respectively and evaluate on the SST2, SearchQA and GSM8K test sets respectively, with results shown in Figure~\ref{pic: gene}(a). Since DGAO has not been trained on the tasks corresponding to the test sets, the model tends to be less confident when answering questions from the test sets, making direct comparison of Consistency Rate with SFT models unfair. Therefore, we choose to compare the ratio of Consistency Rate to Accuracy, which represents the proportion of order-stable outputs the model achieves under a fair accuracy baseline. The results show that compared to naive SFT on the training sets corresponding to the test datasets, DGAO-trained models also achieve improved order stability on other datasets, indicating that DGAO's order stability can generalize to tasks beyond the training tasks. Additionally, we also evaluate whether the model still has generalizable order stability under longer contexts. As shown in Figure~\ref{pic: gene}(b), we test the performance of models trained with different methods on the SQuAD v2 dataset containing 32 documents in the context. The results show that the DGAO-trained model demonstrates consistent accuracy and order stability. Compared to PAFT, DGAO improves accuracy by 26.41\% and Consistency Rate by 16.68\%, indicating that DGAO's order stability can generalize to tasks with more contextual data than the training data. More experiments on datasets and longer context can be found in Appendix~\ref{ap:generalization}.

\section{Acknowledgment} This work is supported by the National Key R\&D Program of China [2023YFC3304903].

\section{Conclusion}
In this paper, we propose DGAO that simultaneously improves model accuracy and order stability. Extensive experiments on multiple datasets demonstrate that DGAO can make models more inclined toward order fairness.


\section{Limitations}
In our experiments, due to computational cost constraints, we primarily conduct experiments on LLMs with 3B-14B parameters. In future work, we will further validate the scalability of DGAO on models with larger parameter counts. Additionally, in our experiments, DGAO uses a fixed number of random order variants. While this choice balances computational efficiency and coverage, for contexts containing many elements, a limited number of variants may not adequately cover all possible ordering patterns. Although the number of order variants learned by the model can be supplemented by increasing the sampling size, training epochs, or training sample size, this would incur additional computational overhead and data collection costs. In future work, we will explore adaptive sampling strategies or difficulty-based order selection methods.

\bibliography{custom}

\clearpage
\appendix
\section{Theoretical Motivation}
\label{Appendix: Theoretical Motivation}
To mitigate order sensitivity, we aim to optimize the policy $\pi_\theta$ so that it consistently generates correct answers under different permutations. Let $y^*$ be the ground truth answer. We first theoretically analyze the limitations of existing strategies in addressing this problem. For standard supervised fine-tuning on augmented data (PAFT~\cite{Zhang2024PositionAwarePE}), it minimizes the negative log-likelihood:

{
\begin{equation}
\mathcal{L}_{\text{PAFT}}(\theta) = - \mathbb{E}_{\sigma \sim \Sigma_n} [\log \pi_\theta(y^* | x_\sigma)].
\end{equation}
}

While this encourages the model to output $y^*$ for all permutations, it treats each permutation $x_\sigma$ as an independent sample. It does not explicitly penalize the difference between distributions $\pi_\theta(\cdot | x_\sigma)$ and $\pi_\theta(\cdot | x_{\sigma'})$. As observed in Table~\ref{tab:intro}, forcing the model to memorize $y^*$ corresponding to different ordered permutations $x_\sigma$ without constraints often leads to overfitting or catastrophic forgetting, harming general accuracy. 

Similarly, standard reinforcement learning approaches maximize the expected reward:

{
\begin{equation}
\mathcal{J}(\theta) = \mathbb{E}_{\sigma \sim \Sigma_n} \mathbb{E}_{y \sim \pi_\theta(\cdot|x_\sigma)} [R(y, y^*)].
\end{equation}
}

However, the traditional reward $R(y, y^*)$ is only based on correctness, so gradient updates independently increase the probability of correct samples. Similar to standard SFT and PAFT, it likewise lacks a mechanism to explicitly reward policy consistency across the permutation group $\{q_{\pi_1}, \ldots, q_{\pi_N}\}$.

\section{Experimental Setup Details}
\label{ap:exp}
All experiments use 8 NVIDIA A800-SXM4 80G GPUs, with memory size of 1.5 TB, operating system Ubuntu 18.04.6, Python version 3.10.18, torch version 2.6.0, CUDA version 12.0, transformers version 4.53.3.

For SFT and PAFT, we set global batch size to 8, learning rate to 2e-5, and train for 3 epochs.

For DGAO, global batch size to 8, we set learning rate to 2e-5, random order number to 8, i.e., 8 order variants of each original query are allocated on each device. Temperature and top p sampling are both set to 1.0. $\epsilon_A$ is set to 1e-8. $\epsilon$ is set to 0.2. We train DGAO for 3 epochs.

For PPO and GRPO, we set global batch size to 8, learning rate to 2e-5, training epoch to 3, temperature and top p sampling are both set to 1.0, $\epsilon$ is set to 0.2. GRPO's generation number is set to 8.

For the above training, the Adam optimizer parameters are $\beta_1=0.9$, $\beta_2=0.999$, $\epsilon=1e-8$.

All evaluation results are based on the average of 3 independent runs and use greedy decoding.

\section{Supplementary Generalization Experiments}
\label{ap:generalization}
Table~\ref{tab:longer_context} shows the results of evaluating the model's generalization ability for order stability on longer contexts. On the 32-shot and 64-shot SQuAD v2 datasets, DGAO achieves higher accuracy and order stability, while PAFT reduces accuracy by an average of 23.7\%. On the SST2 dataset, although PAFT improves the Consistency Rate, the Overconfidence Rate increases. This indicates that PAFT generates consistently incorrect answers in pursuit of high answer consistency. Overall, models trained with DGAO also demonstrate generalized order stability and accuracy improvements on longer contexts.

\begin{table}[ht]
\resizebox{0.48\textwidth}{!}{
\begin{tabular}{llllccc}
\toprule
\textbf{Task} & \textbf{Dataset} & \textbf{Method} & \textbf{Accuracy $\uparrow$} & \textbf{Consistency Rate $\uparrow$} & \textbf{Overconfidence Rate $\downarrow$} \\
\midrule
\multirow{6}{*}{\textbf{32-shot}} & \multirow{3}{*}{SQuAD v2} & SFT & 60.61 & 77.24 & 21.05 \\
& & PAFT & 37.48 (-23.13) & 61.89 (-15.35) & 28.70 (+7.65) \\
& & \textbf{DGAO} & \textbf{63.89 (+3.28)} & \textbf{78.57 (+1.33)} & \textbf{19.24 (-1.81)} \\
\cmidrule{2-6}
& \multirow{3}{*}{SST2} & SFT & 57.10 & 70.77 & 23.87 \\
& & PAFT & 55.78 (-1.32) & \textbf{81.42 (+10.65)} & 34.82 (+10.95) \\
& & \textbf{DGAO} & \textbf{57.91 (+0.81)} & 70.84 (+0.07) & \textbf{23.60 (-0.27)} \\
\midrule
\multirow{6}{*}{\textbf{64-shot}} & \multirow{3}{*}{SQuAD v2} & SFT & 49.25 & 67.50 & 23.41 \\
& & PAFT & 24.99 (-24.26) & 50.14 (-17.36) & 29.65 (+6.24) \\
& & \textbf{DGAO} & \textbf{52.75 (+3.50)} & \textbf{68.33 (+0.83)} & \textbf{20.89 (-2.52)} \\
\cmidrule{2-6}
& \multirow{3}{*}{SST2} & SFT & 56.21 & 69.11 & 23.08 \\
& & PAFT & 55.95 (-0.26) & \textbf{73.81 (+4.70)} & 27.28 (+4.20) \\
& & \textbf{DGAO} & \textbf{56.65 (+0.44)} & 70.26 (+1.15) & \textbf{23.05 (-0.03)} \\
\bottomrule
\end{tabular}
}
\caption{Evaluation results on longer context lengths (using Llama-3.2-3B as the base model).}
\label{tab:longer_context}
\end{table}

\section{Scalability of DGAO on Larger Parameter LLMs}
\label{app:larger scale}

We test the effect of DGAO on larger parameter models than those in the main experiments (Section \ref{mainresults}). As shown in Table \ref{tab:larger_scale}, DGAO still demonstrates consistent improvements in Accuracy and Consistency Rate on larger parameter models, and reduces the Overconfidence Rate, which aligns with the results we observe in the main experiments. This indicates that DGAO is scalable and possesses the capability to mitigate order sensitivity on LLMs with larger parameter counts as well.

\begin{table}[htb]
\resizebox{.49\textwidth}{!}{
\begin{tabular}{llccc}
\toprule
\textbf{Model} & \textbf{Dataset} & \textbf{Accuracy $\uparrow$} & \textbf{Consistency Rate $\uparrow$} & \textbf{Overconfidence Rate $\downarrow$} \\
\midrule
\multirow{15}{*}{Qwen-2.5-14B} 
& SST2-SFT & 94.15 & 97.05 & 7.52 \\
& SST2-PAFT & 92.23 (-1.92) & 98.88 (+1.83) & 7.75 (+0.23) \\
& \textbf{SST2-DGAO} & \textbf{94.27 (+0.12)} & \textbf{98.91 (+1.86)} & \textbf{7.41 (-0.11)} \\
\cmidrule{2-5}
& SQuAD v2-SFT & 86.92 & 94.90 & 11.16 \\
& SQuAD v2-PAFT & 83.68 (-3.24) & \textbf{95.41 (+0.51)} & 13.52 (+2.36) \\
& \textbf{SQuAD v2-DGAO} & \textbf{87.51 (+0.59)} & 95.28 (+0.38) & \textbf{10.42 (-0.74)} \\
\cmidrule{2-5}
& SearchQA-SFT & 67.90 & 87.46 & 20.65 \\
& SearchQA-PAFT & 69.03 (+1.13) & \textbf{92.71 (+5.25)} & 24.08 (+3.43) \\
& \textbf{SearchQA-DGAO} & \textbf{71.22 (+3.32)} & 89.82 (+2.36) & \textbf{19.42 (-1.23)} \\
\cmidrule{2-5}
& GSM8k-SFT & 57.90 & 69.44 & 28.57 \\
& GSM8k-PAFT & 49.43 (-8.47) & 69.91 (+0.47) & 36.73 (+8.16) \\
& \textbf{GSM8k-DGAO} & \textbf{58.19 (+0.29)} & \textbf{70.02 (+0.58)} & \textbf{28.27 (-0.30)} \\
\cmidrule{2-5}
& CM17k-SFT & 38.87 & 79.97 & 51.27 \\
& CM17k-PAFT & 29.76 (-9.11) & \textbf{85.08 (+5.11)} & 64.58 (+13.31) \\
& \textbf{CM17k-DGAO} & \textbf{39.75 (+0.88)} & 81.73 (+1.76) & \textbf{47.36 (-3.91)} \\
\midrule
\multirow{15}{*}{Llama-3.2-11B}
& SST2-SFT & 90.23 & 99.18 & 9.87 \\
& SST2-PAFT & 88.47 (-1.76) & 99.24 (+0.06) & 11.43 (+1.56) \\
& \textbf{SST2-DGAO} & \textbf{91.31 (+1.08)} & \textbf{99.27 (+0.09)} & \textbf{8.41 (-1.46)} \\
\cmidrule{2-5}
& SQuAD v2-SFT & 75.65 & 90.95 & 18.61 \\
& SQuAD v2-PAFT & 60.84 (-14.81) & 85.12 (-5.83) & 28.76 (+10.15) \\
& \textbf{SQuAD v2-DGAO} & \textbf{77.28 (+1.63)} & \textbf{91.46 (+0.51)} & \textbf{17.71 (-0.90)} \\
\cmidrule{2-5}
& SearchQA-SFT & 58.44 & 83.16 & 25.96 \\
& SearchQA-PAFT & 56.18 (-2.26) & 85.27 (+2.11) & 31.28 (+5.32) \\
& \textbf{SearchQA-DGAO} & \textbf{62.13 (+3.69)} & \textbf{86.54 (+3.38)} & \textbf{25.52 (-0.44)} \\
\cmidrule{2-5}
& GSM8k-SFT & 32.14 & 58.77 & 37.65 \\
& GSM8k-PAFT & 23.96 (-8.18) & 59.84 (+1.07) & 46.51 (+8.86) \\
& \textbf{GSM8k-DGAO} & \textbf{32.49 (+0.35)} & \textbf{60.35 (+1.58)} & \textbf{35.02 (-2.63)} \\
\cmidrule{2-5}
& CM17k-SFT & 21.80 & 72.38 & 56.85 \\
& CM17k-PAFT & 17.03 (-4.77) & \textbf{75.92 (+3.54)} & 65.62 (+8.77) \\
& \textbf{CM17k-DGAO} & \textbf{21.87 (+0.07)} & 72.56 (+0.18) & \textbf{51.58 (-5.27)} \\
\bottomrule
\end{tabular}
}
\caption{Accuracy and order stability metrics of LLMs with different training methods. $+/-$ indicates improvement or degradation relative to the SFT baseline. $\uparrow$ indicates that higher values of this metric represent better model accuracy or stronger order stability, $\downarrow$ indicates that lower values of this metric represent lower model hallucination.}
\label{tab:larger_scale}
\end{table}

\section{DGAO Algorithm Pseudocode}
\label{ap:dgao}
\begin{algorithm}
\scriptsize
\caption{Dual Group Advantage Optimization (DGAO)}
\label{alg:dgao}
\begin{algorithmic}[1]
\Require Policy model $\pi_\theta$, reference model $\pi_{ref}$, old policy $\pi_{old}$
\Require Batch of $M$ queries, order variations $N$ per query, hyperparameter $\alpha \in [0,1]$, Clipping parameter $\epsilon$, KL penalty $\beta$, small constant $\epsilon_A$

\For{each training iteration}
    \State \textbf{Sample Generation:}
    \For{$j = 1$ to $M$}
        \State Generate $N$ different order versions: $\{q_{j,1}, q_{j,2}, \ldots, q_{j,N}\}$
        \For{$i = 1$ to $N$}
            \State Generate response: $y_{j,i} \sim \pi_\theta(\cdot|q_{j,i})$
            \State Calculate reward: $R(y_{j,i}) = \begin{cases} 1 & \text{if correct} \\ 0 & \text{if incorrect} \end{cases}$
        \EndFor
    \EndFor
    
    \State \textbf{Advantage Calculation:}
    \For{$j = 1$ to $M$}
        \State Calculate group reward: $R_{\text{group}}(j) = \frac{1}{N} \sum_{i=1}^{N} R(y_{j,i})$
    \EndFor
    
    \State Calculate global baseline: $b = \frac{1}{M} \sum_{j=1}^{M} R_{\text{group}}(j)$
    
    \For{$j = 1$ to $M$}
        \State Calculate inter-group advantage: $A_{\text{group}}(j) = R_{\text{group}}(j) - b$
        \For{$i = 1$ to $N$}
            \State Calculate intra-group advantage: $A_{\text{ind}}(y_{j,i}) = R(y_{j,i}) - R_{\text{group}}(j)$
            \State Calculate hybrid advantage: 
            \State \hspace{2em} $A_\text{hybrid}(y_{j,i}) = \alpha \cdot A_\text{group}(j) + (1-\alpha) \cdot A_\text{ind}(y_{j,i})$
        \EndFor
    \EndFor
    
    \State \textbf{Advantage Normalization:}
    \State Calculate batch statistics: $\mu_A = \text{mean}(A_\text{hybrid}), \sigma_A = \text{std}(A_\text{hybrid})$
    \For{$j = 1$ to $M$, $i = 1$ to $N$}
        \State $A_\text{final}(y_{j,i}) = \frac{A_\text{hybrid}(y_{j,i}) - \mu_A}{\sigma_A + \epsilon_A}$
    \EndFor
    
    \State \textbf{Policy Update:}
    \State Calculate probability ratios: $r_{j,i} = \frac{\pi_\theta(y_{j,i}|q_{j,i})}{\pi_{old}(y_{j,i}|q_{j,i})}$
    \State Calculate clipped objective:
    \State \hspace{2em} $L_{\text{clip}}(y_{j,i}) = \min(r_{j,i} A_\text{final}(y_{j,i}), \text{clip}(r_{j,i}, 1-\epsilon, 1+\epsilon) A_\text{final}(y_{j,i}))$
    \State Calculate KL penalty: $L_{\text{KL}} = \mathbb{D}_{KL}(\pi_\theta || \pi_{ref})$
    \State Calculate total loss: 
    \State \hspace{2em} $\mathcal{L}_{\text{DGAO}}(\theta) = \frac{1}{MN} \sum_{j=1}^{M} \sum_{i=1}^{N} L_{\text{clip}}(y_{j,i}) - \beta L_{\text{KL}}$
    \State Update policy parameters: $\theta \leftarrow \theta - \nabla_\theta \mathcal{L}_{\text{DGAO}}(\theta)$
    \State Update old policy: $\pi_{old} \leftarrow \pi_\theta$
\EndFor

\State \Return Optimized policy model $\pi_\theta$
\end{algorithmic}
\end{algorithm}

\newpage

\section{Hyperparameter \texorpdfstring{$\alpha$}{Alpha} Sweeping Experiments}
To more precisely analyze the impact of hyperparameter $\alpha$ on DGAO, we conduct parameter sweeping experiments on SST2 and SearchQA tasks, as shown in Table~\ref{tab:alpha_sweep}. It is observed that when $\alpha$ is in the range of 0.4 to 0.6, both inter-group advantage and intra-group advantage take effect, achieving the best performance by balancing accuracy and order consistency. When $\alpha \rightarrow 0$, the Consistency Rate (CR) decreases, indicating that weakening the intra-group advantage leads to order instability in the model; when $\alpha \rightarrow 1$, accuracy (Acc) decreases and Overconfidence Rate (OR) increases, indicating that weakening the inter-group advantage leads to reduced accuracy and increased hallucination rate. This demonstrates that the two types of advantage computation in DGAO are complementary and indispensable.
\begin{table}[ht]
\centering
\resizebox{0.48\textwidth}{!}{%
\begin{tabular}{l|ccc|ccc}
\toprule
\multicolumn{1}{c}{\multirow{2}{*}{\textbf{Setting}}} & \multicolumn{3}{c}{\textbf{SST2}} & \multicolumn{3}{c}{\textbf{SearchQA}} \\
\cmidrule{2-7}
\multicolumn{1}{c}{} & \multicolumn{1}{c}{Accuracy $\uparrow$} & \multicolumn{1}{c}{CR $\uparrow$} & \multicolumn{1}{c}{OR $\downarrow$} & \multicolumn{1}{c}{Accuracy $\uparrow$} & \multicolumn{1}{c}{CR $\uparrow$} & \multicolumn{1}{c}{OR $\downarrow$} \\
\midrule
$\alpha=0.0$ & 90.29 & 98.83 & 9.39 & 61.98 & 80.30 & 25.51 \\
$\alpha=0.1$ & 90.46 & 98.90 & 9.15 & 62.12 & 81.50 & 25.48 \\
$\alpha=0.2$ & 90.61 & 98.98 & 8.82 & 62.25 & 82.60 & 25.46 \\
$\alpha=0.3$ & 90.52 & 99.05 & 9.00 & 62.38 & 82.50 & 25.40 \\
$\alpha=0.4$ & 90.59 & 99.10 & 8.90 & 62.46 & 84.20 & 25.44 \\
$\alpha=0.5$ & \textbf{90.64} & \textbf{99.14} & \textbf{8.80} & \textbf{62.51} & \textbf{84.72} & \textbf{25.43} \\
$\alpha=0.6$ & 90.72 & 99.18 & 8.72 & 62.45 & 85.10 & 25.65 \\
$\alpha=0.7$ & 90.40 & 99.25 & 9.05 & 61.20 & 86.80 & 25.85 \\
$\alpha=0.8$ & 89.78 & 99.30 & 9.95 & 59.80 & 87.50 & 26.40 \\
$\alpha=0.9$ & 89.80 & 99.00 & 9.80 & 58.50 & 84.10 & 26.90 \\
$\alpha=1.0$ & 89.44 & 99.04 & 10.08 & 57.82 & 80.39 & 26.19 \\
\bottomrule
\end{tabular}%
}
\caption{Hyperparameter $\alpha$ sweeping experiments (using Llama-3.2-3B as the base model).}
\label{tab:alpha_sweep}
\end{table}

\begin{figure*}[t]
\centering
\includegraphics[width=0.98\textwidth]{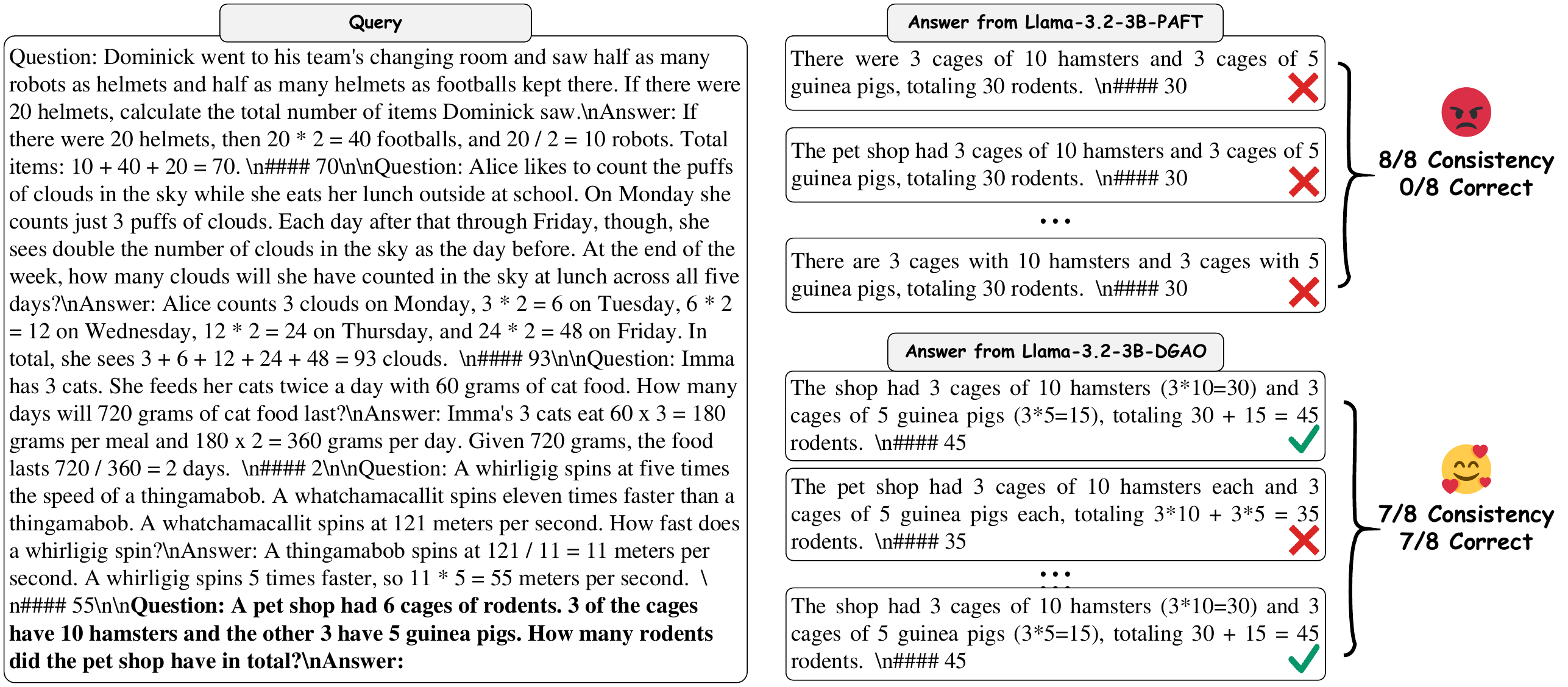} 
\caption{Comparison examples of PAFT and DGAO test results.}
\label{pic:exp_badcase}
\end{figure*}

\section{Supplementary Results}
\label{ap:supplementary}
Figure~\ref{pic:exp_badcase} shows a bad case of the PAFT-trained model evaluated on the GSM8K dataset. For 8 order variants of a query, PAFT provides consistently incorrect answers, indicating that PAFT generates consistently wrong responses in pursuit of consistency. While DGAO has 1 inconsistent generation, the overall accuracy reaches 7/8. This shows that DGAO balances the order stability and accuracy of generation.

\begin{figure*}[t]
\centering
\includegraphics[width=0.95\textwidth]{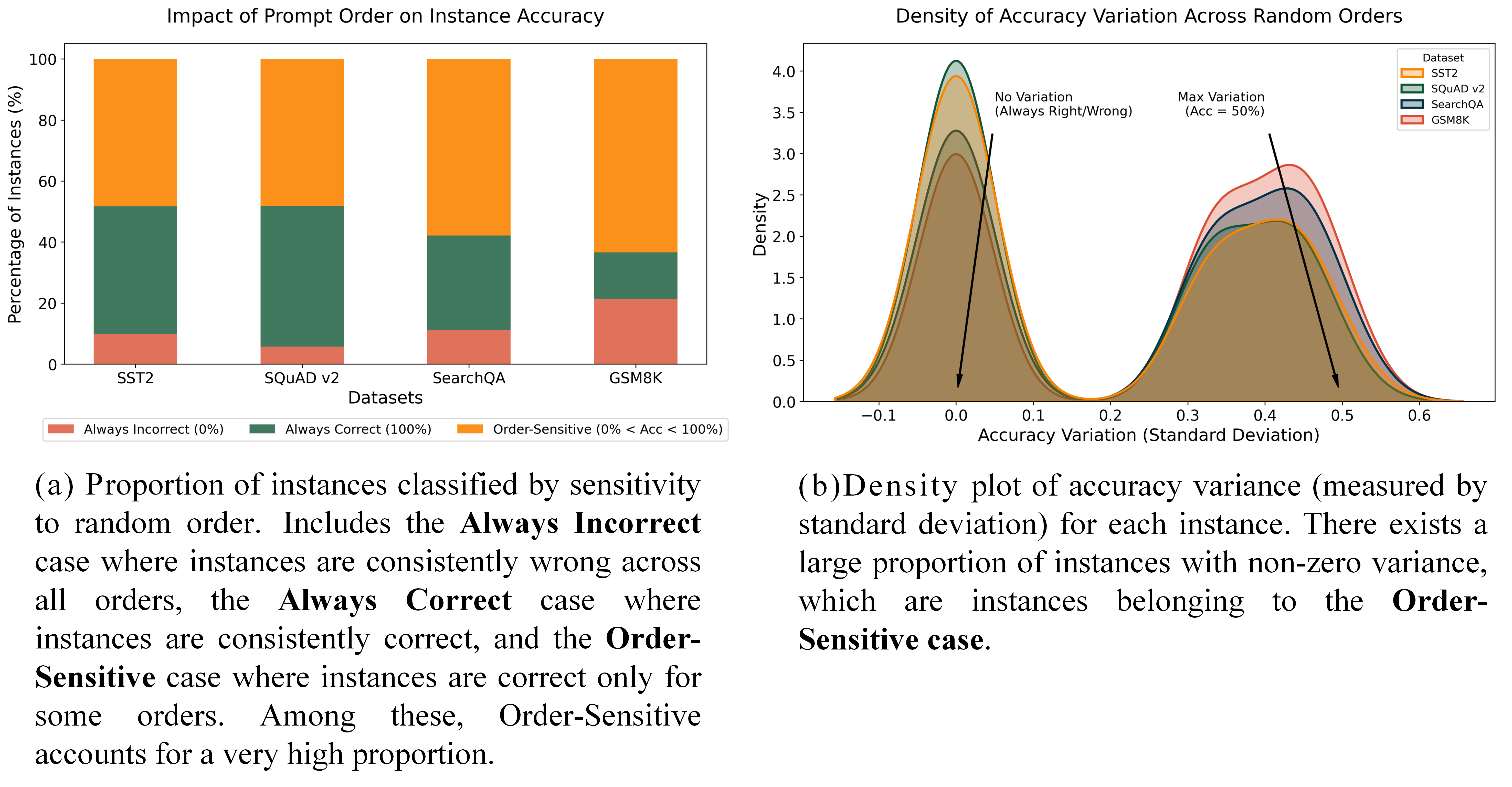} 
\caption{Impact of input order on the performance of Qwen-2.5-3B across 8 random orders.}
\label{Qwen-2.5-3B}
\end{figure*}


\begin{figure*}[t]
\centering
\includegraphics[width=0.95\textwidth]{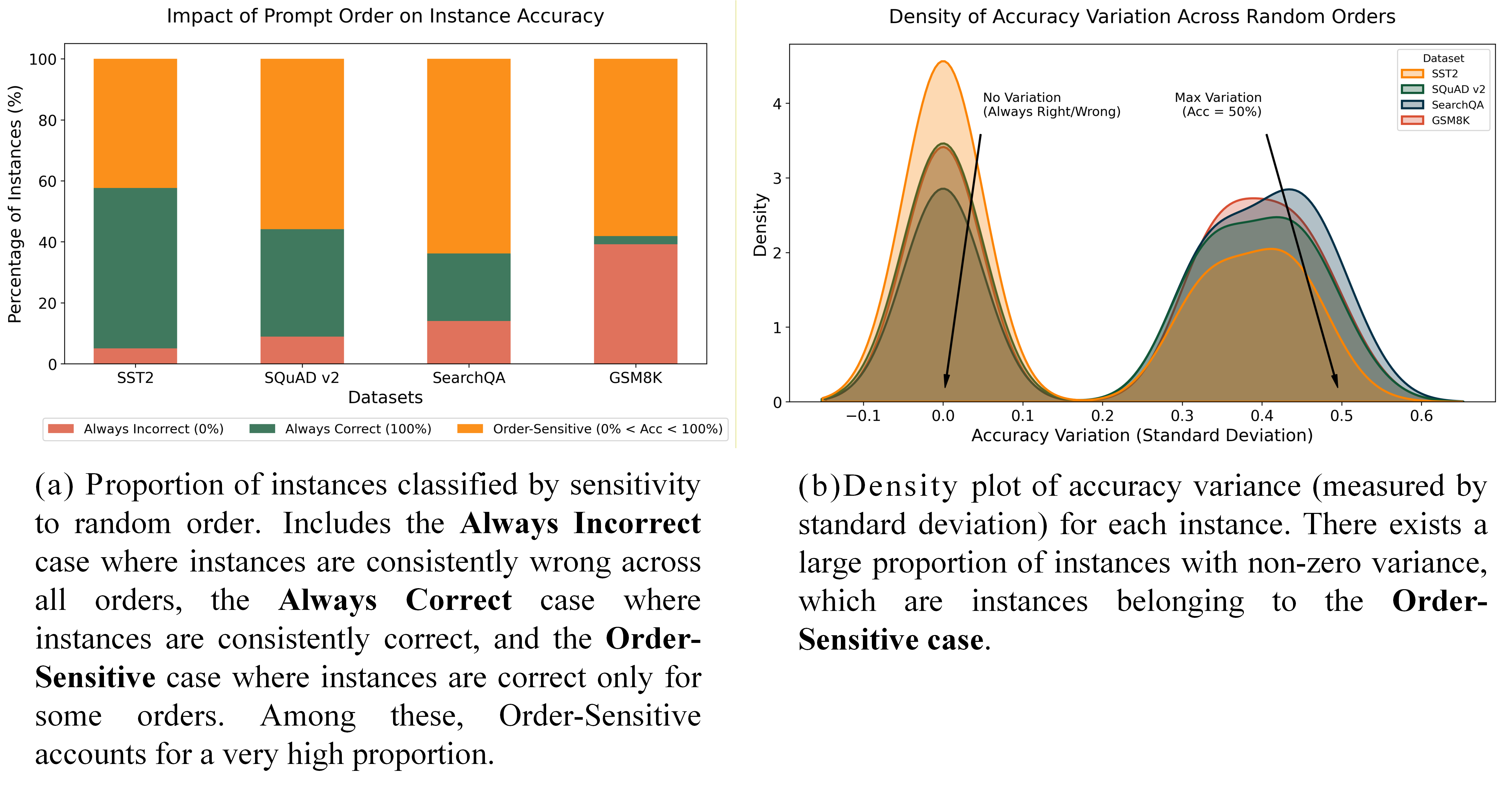} 
\caption{Impact of input order on the performance of Gemma-3-4B across 8 random orders.}
\label{Gemma-3-4B}
\end{figure*}

Figure~\ref{Qwen-2.5-3B} and Figure~\ref{Gemma-3-4B} are supplementary to Figure~\ref{pic: orderbias} in the main text. The results indicate that all three models are order-sensitive.

\section{Data Construction}
\label{ap:data}
\begin{figure}[ht]
\centering
\includegraphics[width=0.48\textwidth]{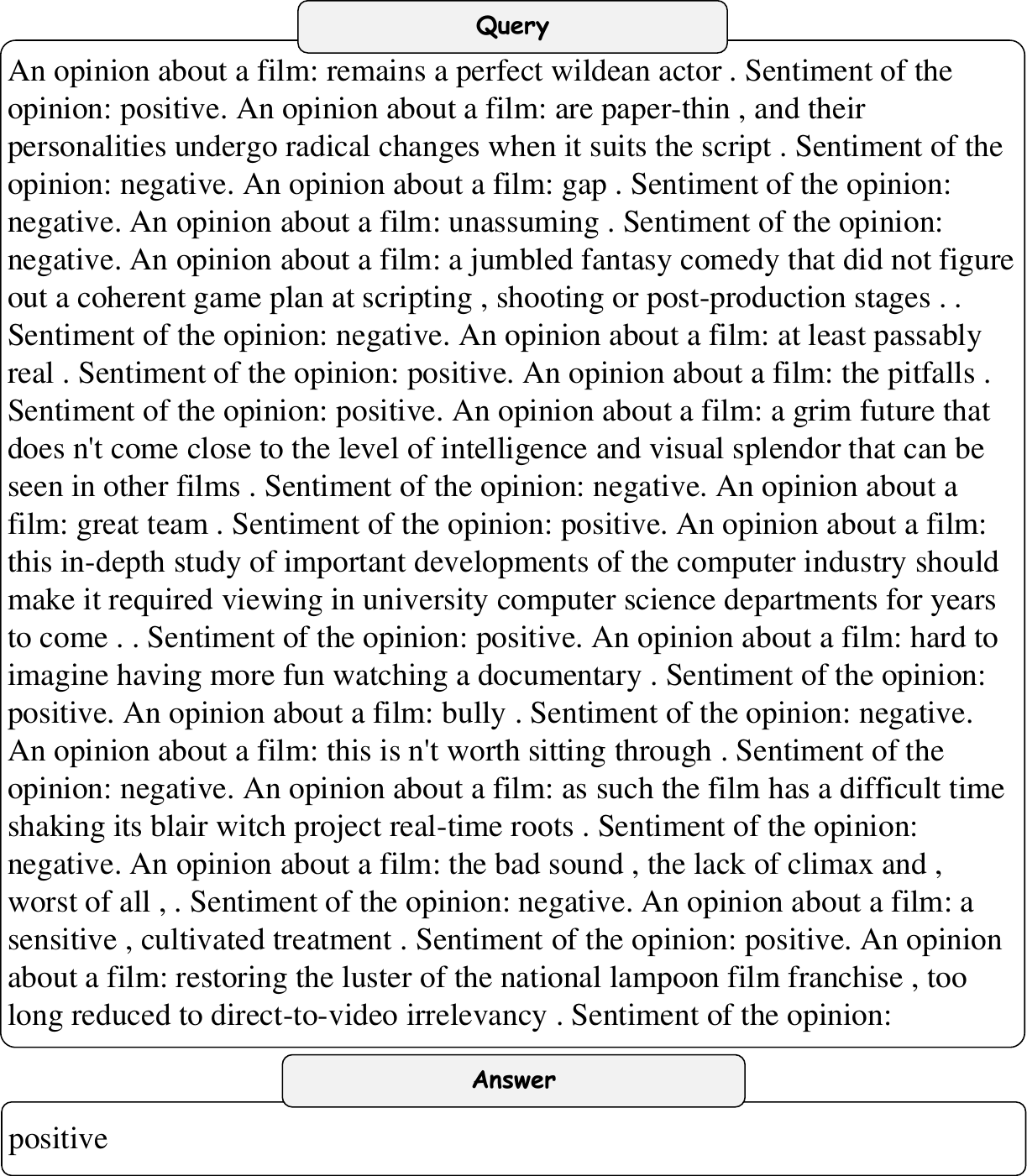} 
\caption{Example of SST2 dataset.}
\label{exp_sst2}
\end{figure}

\begin{figure}[ht]
\centering
\includegraphics[width=0.48\textwidth]{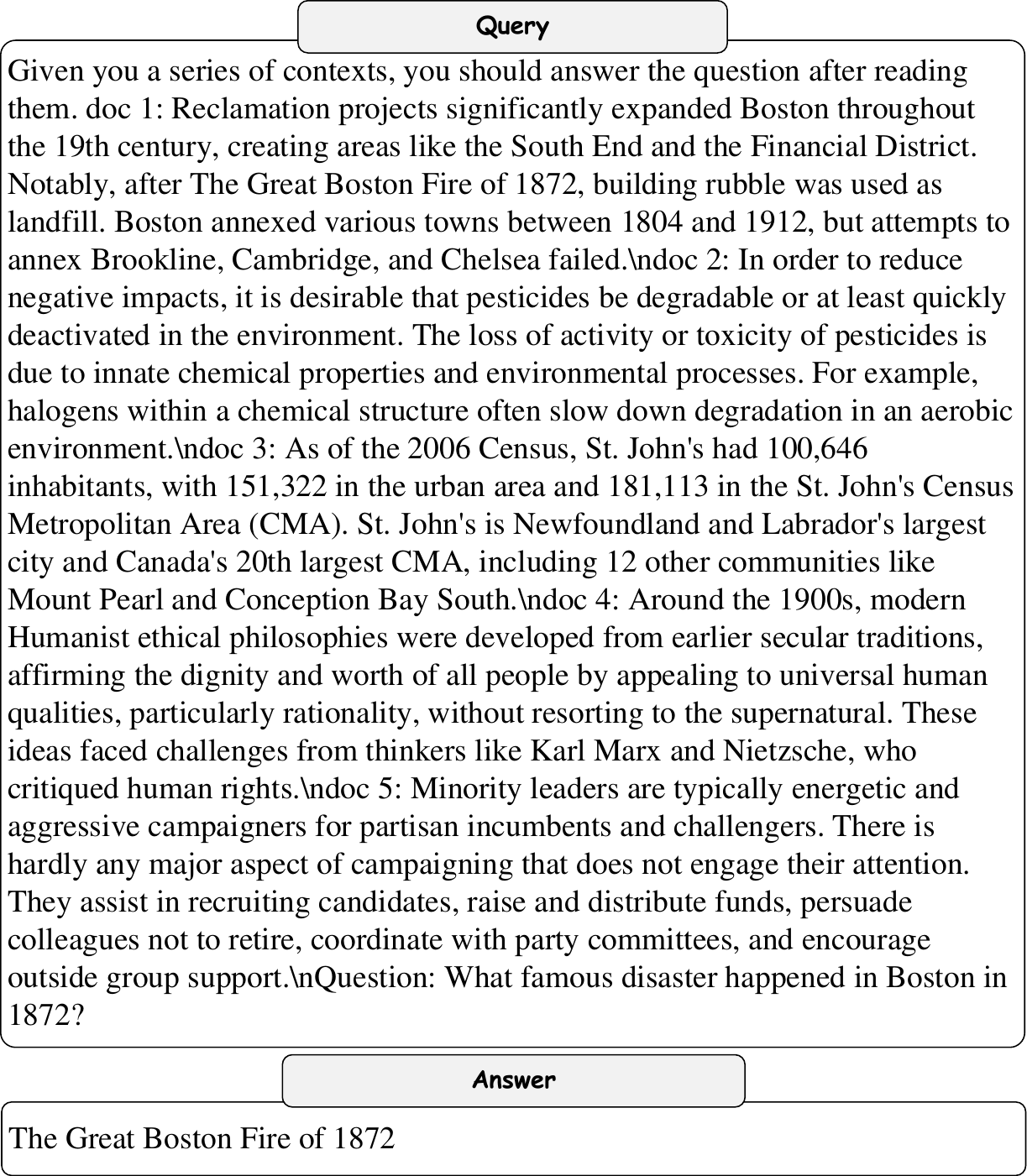} 
\caption{Example of SQuAD v2 dataset.}
\label{exp_squad}
\end{figure}

\begin{figure}[ht]
\centering
\includegraphics[width=0.48\textwidth]{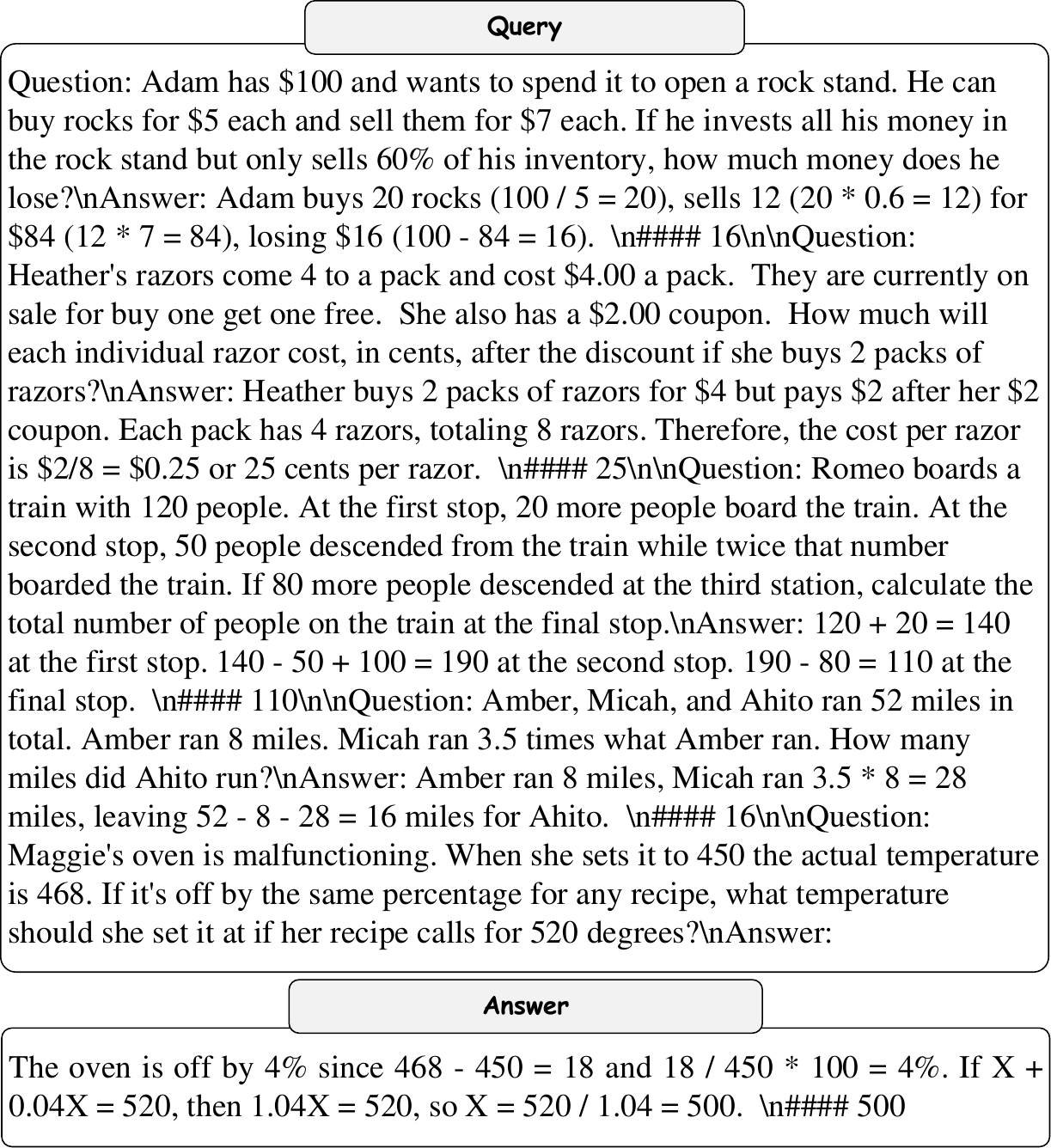} 
\caption{Example of GSM8K dataset.}
\label{exp_gsm8k}
\end{figure}

Figures~\ref{exp_sst2}, \ref{exp_squad}, and \ref{exp_gsm8k} respectively show construction examples of the SST2, SQuAD v2, and GSM8k datasets.

\section{Computational Cost of DGAO}
\label{ap:cost}
As an online reinforcement learning method, DGAO has comparable training overhead to GRPO, both of which are slightly higher than SFT. Taking the SearchQA task as an example, training the Qwen-3B model for 3 epochs using DGAO, GRPO, and SFT requires 0.85, 0.81, and 0.22 GPU hours respectively on 8 A800 GPUs with 80G memory. This is mainly because DGAO and GRPO require multiple sampling rounds and advantage computation, which is unavoidable in online reinforcement learning. Considering the capability improvements that DGAO brings, this additional overhead is acceptable.

During inference, the model trained with DGAO has the same inference overhead and time consumption as the model after SFT, because DGAO neither introduces additional modules nor expands the input during inference. Compared to methods that use dataset-based search to find optimal or suboptimal arrangement orders~\cite{Lu2021FantasticallyOP,sorensen2022information,Guo2024WhatMA}, DGAO also has the advantage of inference time.

\end{document}